\documentclass{article} %
\usepackage{main_conference,times}

\usepackage{amsmath,amsfonts,bm}

\def\eqref#1{equation~\ref{#1}}

\def\1{\bm{1}}

\DeclareMathAlphabet{\mathsfit}{\encodingdefault}{\sfdefault}{m}{sl}
\SetMathAlphabet{\mathsfit}{bold}{\encodingdefault}{\sfdefault}{bx}{n}

\newcommand{\E}{\mathbb{E}}

\newcommand{\R}{\mathbb{R}}

\usepackage{hyperref}
\usepackage{xcolor}
\usepackage{url}

\usepackage[shortlabels]{enumitem}
\usepackage{wrapfig,lipsum,booktabs}
\usepackage{graphicx}
\usepackage[font=footnotesize,labelfont=bf]{caption}
\usepackage{amsmath} 
\usepackage{url}
\usepackage{seqsplit}
\usepackage{hyperref}
\hypersetup{
	colorlinks=true,
	linkcolor=orange,
	filecolor=magenta,      
	urlcolor=orange,
	citecolor=orange,
}
\usepackage{colortbl}
\usepackage[normalem]{ulem}

\usepackage{algorithm}
\usepackage{algpseudocode}
\usepackage{amsmath,amssymb,amsthm}
\usepackage{multirow}
\usepackage{wrapfig}
\usepackage{longtable}
\newcommand{\N}{\mathcal{N}}
\newcommand{\U}{\mathcal{U}}
\newcommand{\Loss}{\mathcal{L}}

\newcommand{\azero}{\mathbf{a}_0}

\newcommand{\vvideo}{v_\theta^{\text{video}}}
\newcommand{\vaction}{v_\phi^{\text{action}}}

\usepackage{titlesec}
\titlespacing\section{0pt}{3.3pt plus 4pt minus 2pt}{3.3pt plus 2pt minus 2pt}
\titlespacing\subsection{0pt}{1.5pt plus 4pt minus 2pt}{1.5pt plus 2pt minus 2pt}

\usepackage{listings}
\title{\Large DiT4DiT: Jointly Modeling Video Dynamics and Actions for Generalizable Robot Control}

\newcommand{\website}{\url{https://dit4dit.github.io/}}

\author{
\hspace{1.2cm}\textbf{Teli Ma}$^{1,2}$ \hspace{0.18cm}
\textbf{Jia Zheng}$^{1,2}$ \hspace{0.18cm}
\textbf{Zifan Wang}$^{1,2}$ \hspace{0.18cm}
\textbf{Chunli Jiang}$^{1}$ \hspace{0.18cm}
\textbf{Andy Cui}$^{1}$ \\[0.8ex]
\hspace{4.4cm}\textbf{Junwei Liang}$^{2,3,*}$ \hspace{0.18cm}
\textbf{Shuo Yang}$^{1,*}$ 
\\[1.2ex]
\hspace{3.2cm}$^{1}$Mondo Robotics \hspace{0.18cm} $^{2}$HKUST(GZ) \hspace{0.18cm} $^{3}$HKUST\\[0.18cm]
\hspace{4.2cm}$^{*}$Corresponding author, Co-advising \\
\vspace{-20pt}
}

\iclrfinalcopy %
\begin{document}

\maketitle
\lhead{}  %

\begin{abstract}
Vision-Language-Action (VLA) models have emerged as a promising paradigm for robot learning, but their representations are still largely inherited from static image-text pretraining, leaving physical dynamics to be learned from comparatively limited action data. Generative video models, by contrast, encode rich spatiotemporal structure and implicit physics, making them a compelling foundation for robotic manipulation. But their potentials are not fully explored in the literature. To bridge the gap, we introduce DiT4DiT, an end-to-end Video-Action Model that couples a video Diffusion Transformer with an action Diffusion Transformer in a unified cascaded framework. Instead of relying on reconstructed future frames, DiT4DiT extracts intermediate denoising features from the video generation process and uses them as temporally grounded conditions for action prediction. We further propose a dual flow-matching objective with decoupled timesteps and noise scales for video prediction, hidden-state extraction, and action inference, enabling coherent joint training of both modules. Across simulation and real-world benchmarks, DiT4DiT achieves state-of-the-art results, reaching average success rates of 98.6\% on LIBERO and 50.8\% on RoboCasa GR1 while using substantially less training data. On the Unitree G1 robot, it also delivers superior real-world performance and strong zero-shot generalization. Importantly, DiT4DiT improves sample efficiency by over 10$\times$ and speeds up convergence by up to 7$\times$, demonstrating that video generation can serve as an effective scaling proxy for robot policy learning. We release code and models at \website.
\end{abstract}

\section{Introduction}
\label{sec:intro}

Vision-Language-Action (VLA) models~\citep{brohan2023can, brohan2023rt, kim2024openvla, black2024pi_0, intelligence2025pi06, bjorck2025gr00t, gr00tn16}, built upon the success of Vision-Language Models (VLMs)~\citep{achiam2023gpt, touvron2023llama, karamcheti2024prismatic, bai2025qwen3}, have demonstrated remarkable capabilities across a wide range of robotic tasks. Yet most existing VLA systems inherit backbones pretrained primarily on static image-text data, leaving spatiotemporal structure and physical dynamics to be learned only during downstream policy training.
In parallel, video generation models (VGMs)~\citep{wan2025, nvidia2025cosmosworldfoundationmodel, cosmos25, cai2025z} have emerged as a promising alternative: by synthesizing temporally coherent and physically plausible futures video frames, they learn rich motion priors, causal structure, and implicit physical dynamics. This suggests a broader opportunity for robotics: beyond serving as auxiliary models, video generators may provide a strong foundation model backbone for robot control.

\begin{figure}[htbp]
    \centering
    \includegraphics[width=1.0\linewidth]{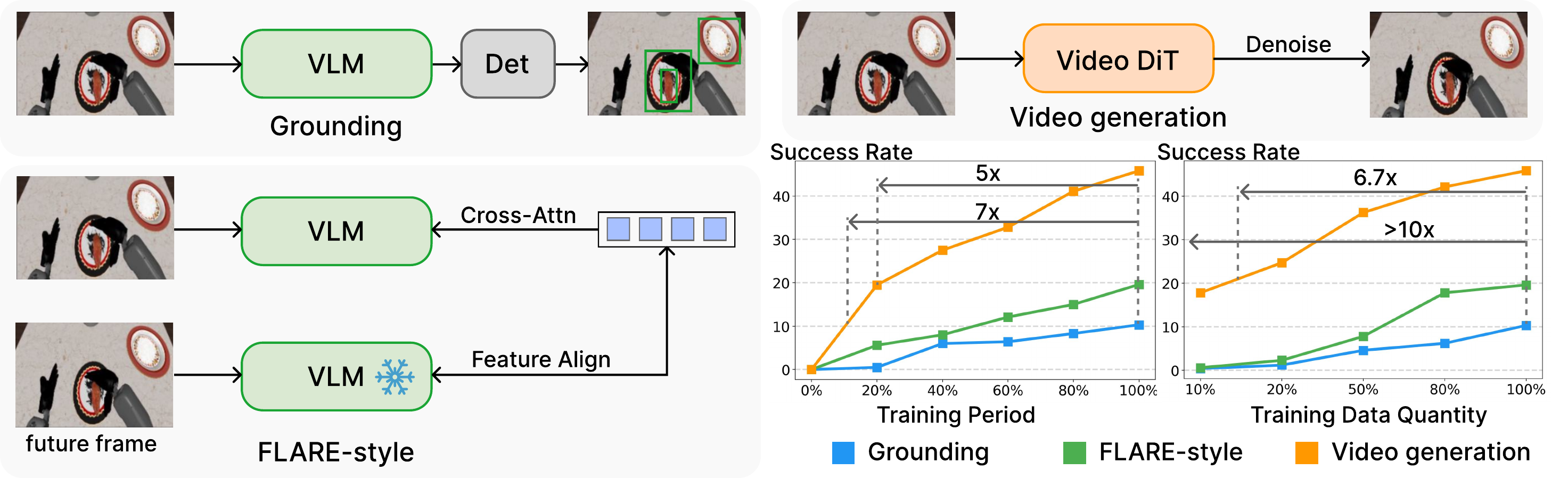}
    \vspace{-15pt}
    \caption{\textbf{Proxy objectives for scalable robot policy learning.} \textbf{Left:} Comparison of three representative training paradigms: \textit{Grounding} (object-level semantic alignment), \textit{FLARE-style}~\citep{zheng2025flare} latent modeling (VLM-to-future-frame feature prediction), and \textit{Video generation} (learning physically plausible future dynamics). \textbf{Right:} Video generation serves as the strongest scaling proxy, yielding higher sample efficiency (up to $>10\times$), faster convergence (up to $7\times$), and more favorable scaling trends across data regimes, with consistently better downstream manipulation success than semantic-centric baselines. All results are reported as the average success rate over 24 tasks in the RoboCasa-GR1 tabletop benchmark~\citep{ nasiriany2024robocasa, bjorck2025gr00t}.}
    \label{fig:intro}
    \vspace{-30pt}
\end{figure}

Recent works~\citep{unifolm-wma-0,liang2025video, feng2025vidar, liao2025genie, wang2025latent, li2025unified, bi2025motus, cosmospolicy, pai2025mimic} have begun exploring this direction, typically by using video models to synthesize additional training data or by extracting latent representations to train inverse dynamics models for action prediction. While encouraging, these approaches are often multi-stage rather than end-to-end, making control indirect and leaving open the central question of how video generative models should be integrated to serve as a principled backbone for policy learning. In this work, we take a step toward that goal by answering two questions: (1) can video generation itself serve as an effective training objective for robust action policies? and (2) how should the spatiotemporal representations learned by video models be extracted and coupled with action generation?

We first examine whether video generation can serve as an effective proxy objective for policy learning. The strong dependence on action-labeled data has long constrained the scaling of VLA models. 
Prior attempts to leverage visual supervision through auxiliary tasks (e.g., grounding and VLM-centric latent feature modeling) are often sample-inefficient. For instance, methods like FLARE~\citep{zheng2025flare} attempt to align current-future representations with pre-trained VLMs, but struggle to capture continuous pixel-level physical dynamics. In contrast, we find that video generation is a highly effective unsupervised pre-training signal. As shown in Fig.~\ref{fig:intro}, our video-dynamics objective converges faster and achieves higher final success rates than both \textit{Grounding} and \textit{FLARE-style} baselines.

To this end, we introduce DiT4DiT, a unified end-to-end Video-Action Model (VAM) with a \textbf{dual-DiT} architecture. Unlike prior methods built on visual-language autoregressive backbones, our framework adopts a bidirectional Video Diffusion Transformer (DiT)~\citep{peebles2023scalable}. During denoising, we extract compact latent features from future-frame generation and use them to condition action learning, so the policy is grounded in the generative visual dynamics that govern physical interaction.
To avoid disjoint multi-stage optimization, we further propose a unified joint-training paradigm based on \textbf{dual flow-matching}, which optimizes video and action generation in one framework. The method assigns separate timesteps and noise scales to the two modules, enabling either independent or coupled updates while transferring denoised multi-stage video latents into the action latent space. This design streamlines the training workflow and significantly shortens the convergence cycle.

We evaluate our method extensively across both simulation and real-world settings to demonstrate its efficacy in translating generative physical priors into precise robotic control. As an end-to-end policy, DiT4DiT achieves a new state-of-the-art on both the LIBERO~\citep{liu2024libero} and RoboCasa-GR1~\citep{nasiriany2024robocasa} Tabletop simulation benchmarks (98.6\% and 50.8\% average success rates, respectively). It demonstrates exceptional extended-horizon capabilities on LIBERO, outperforming recent strong VLA models like $\pi_{0.5}$~\citep{intelligence2025pi} and CogVLA~\citep{li2025cogvla}. On the challenging 24-task RoboCasa-GR1 suite, it decisively surpasses highly optimized, pre-trained policies like the GR00T series~\citep{bjorck2025gr00t, gr00tn16} by substantial margins. 
In real-world Unitree G1 deployments, DiT4DiT maintains clear advantages over both pre-trained (GR00T-N1.5~\citep{bjorck2025gr00t}) and parameter-matched baselines. Remarkably, relying on only a single egocentric camera, our framework extracts rich spatial reasoning capabilities, achieving the high accuracy required for precision-critical tasks such as \textit{Arrange Flower} and \textit{Stack Cup}. Furthermore, DiT4DiT exhibits robust zero-shot generalization under severe distribution shifts, successfully adapting to unseen objects, category changes, and quantity variations in both simulation and physical reality.

\section{Related Works}
This work connects advances in generalist robot policies with recent progress in generative world modeling. We therefore review two complementary lines of research: Visual-language-based models and video-generation-based models.

\subsection{Vision-Language-Action Models}
The emergence of Vision-Language-Action (VLA) models has established a transformative paradigm for generalist robot learning. By co-fine-tuning VLMs on robotic trajectories, models such as RT-2 \citep{brohan2023rt}, OpenVLA \citep{kim2024openvla}, UniVLA \citep{bu2025univla}, CogVLA \citep{li2025cogvla}, GR00T~\citep{bjorck2025gr00t,gr00tn16} and the $\pi$ \citep{black2024pi_0, intelligence2025pi} family successfully transfer semantic priors to embodied control. By inheriting the extensive visual and linguistic representations of their backbones, these policies demonstrate remarkable zero-shot generalization to novel instructions and semantic concepts that are otherwise absent from standard robotic datasets.

Despite their impressive semantic proficiency, a critical limitation of current VLAs stems from their foundational architecture: they rely on representations learned almost exclusively from static image-text pairs.  
Consequently, the heavy burden of learning low-level physical interactions and temporal state transitions falls entirely on the downstream robotic fine-tuning phase, which requires thousands of hours of training data. 
In contrast to these static VLA paradigms, our approach is built upon a pre-trained video diffusion model. Having been optimized to predict future frames across internet-scale video datasets, video generative models~\citep{kong2024hunyuanvideo, zheng2024open, cosmos25, nvidia2025cosmosworldfoundationmodel, wan2025} naturally internalize the complex, continuous physical dynamics of the real world. We hypothesize that harnessing these rich, pre-existing spatiotemporal and physical priors offers a fundamentally superior foundation for learning robust, low-level robotic control policies.

\subsection{Video Generation in Robotics}
To overcome the physical blindness of static VLMs, recent research has increasingly turned to generative video models, which naturally encapsulate rich spatiotemporal priors and complex physical dynamics~\citep{hu2024video, ye2024latent, liang2025video, feng2025vidar, liao2025genie, wang2025latent, zhong2025flowvla, cen2025worldvla,  feng2025vidar, bi2025motus, lingbot}. Historically, video prediction in robotics was primarily utilized for visual foresight, enabling model-based planning by ``imagining" future states \citep{finn2017deep, ebert2018visual, yang2023unisim, du2023learning}. 
However, with the advent of high-fidelity diffusion transformers, a new frontier has emerged that directly integrates video generation into policy learning.

A recent line of work \citep{shen2025videovla, li2025unified, bi2025motus, lingbot} has explored projecting both visual dynamics and control signals into a shared latent space. 
These models effectively consolidate versatile capabilities (such as forward simulation and inverse dynamics) into a single learned system. Building upon this trend of explicit unification, Cosmos Policy \citep{cosmospolicy} further simplifies the adaptation by fine-tuning a pre-trained video diffusion model to directly output robot actions and future expected values, encoding them as contiguous latent frames within the native video diffusion process.
The most closely related work, mimic-video \citep{pai2025mimic}, pairs a pre-trained video backbone with a separate flow-matching action decoder and conditions the policy on partially denoised video latents at an intermediate flow time. In contrast, we explore \emph{joint training} of video and action generation, enabling the action model to learn how to extract effective features across different stages of the video generation process, yielding more robust representations.

\section{Validation of Video Generation as a Scaling Proxy}

A core hypothesis of this work is that video generation is an effective proxy task for robot control. 
Hence, we first test it to ensure that the design choices are grounded in empirical evidence.
We conduct a comparative study against two paradigms. 
The first is object-level grounding as~\citep{bjorck2025gr00t}, training the VLM with an auxiliary detection head to drive the VLM to understand "what" and "where" objects are for VLA. 
The other one is the implicit world modeling method based on VLM like FLARE-style. FLARE~\citep{zheng2025flare} attends features from VLM with learnable queries and aligns the queries with latent embeddings of future observations. We abandon the diffusion process of the queries in FLARE to perform FLARE-like pre-training here. 
We use Qwen3-2B~\citep{bai2025qwen3} and Cosmos-Predict2.5-2B~\citep{cosmos25} as the VLM and Video backbone to ensure that the scale of trainable parameters remains consistent.

We validate on 24 tabletop manipulation tasks involving the GR1 humanoid robot~\citep{nasiriany2024robocasa, bjorck2025gr00t} in the RoboCasa simulation.
To more effectively evaluate the efficacy of the proxy task, we decouple the pre-training phase from the downstream training of the action expert across all three experimental settings. 
The VLM and Video backbones are trained on the target dataset in a self-supervised manner (except that the grounding task uses pre-annotated bounding boxes), and then kept frozen during the fine-tuning of the action expert. 
The empirical results (see Fig.~\ref{fig:intro}) showcase the superiority of the video generation objective in training efficiency and scalability. 
The generative proxy task allows the model to converge to high-performance policies much faster (up to $7\times$), capturing essential manipulation cues early in the training process.
Also, it demonstrates a robust scaling behavior: demonstrating significantly higher data efficiency (up to $10\times$) than semantic-centric based methods and maintaining a consistent performance improvement as the data volume increases.
This validates video generation not only as an efficient training task but as a viable scaling proxy for generalizable robot control.

\section{DiT4DiT: Unleashing the Potential of Video Model}

\begin{figure}[]
    \centering
    \vspace{-10pt}
    \includegraphics[width=0.9\linewidth]{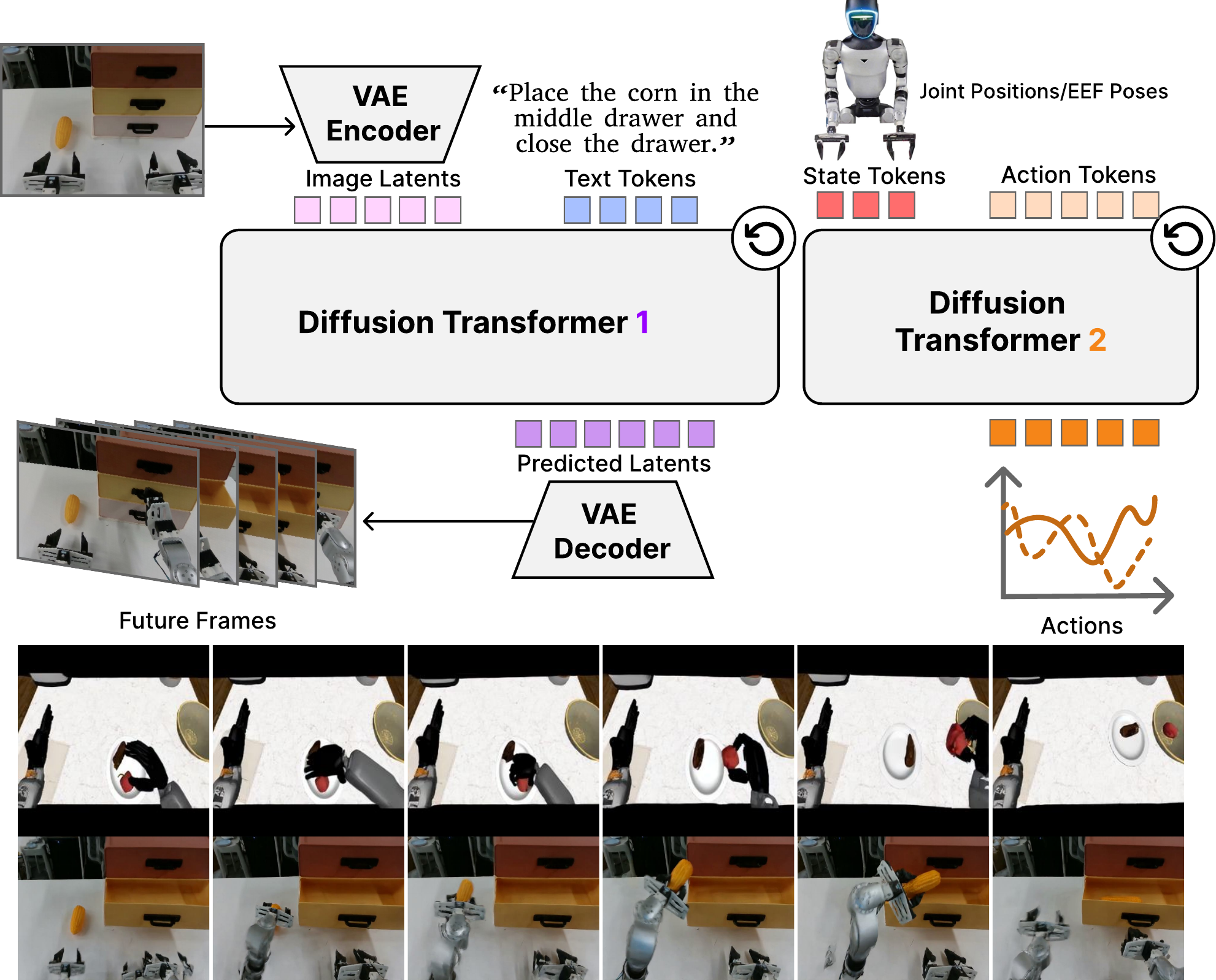}
    \caption{\textbf{Overview of the proposed DiT4DiT framework.} \textbf{Top}: Given the current observation and language goal, the video DiT predicts future dynamics and exposes intermediate generative features at the specific flow timestep; these features condition the action DiT to infer control trajectories. The two models are jointly optimized with a dual flow-matching objective for video generation and action prediction. \textbf{Below}: Generated visual plans via video DiT (More examples are shown in Fig.~\ref{fig:video_gen}).}
    \label{fig:pipe}
    \vspace{-15pt}
\end{figure}

This section details DiT4DiT, an integrated Video-Action Model (VAM) designed for the joint optimization of Video and Action DiTs. By employing a dual flow-matching objective, our framework concurrently refines video synthesis and action prediction. This synergy allows the action policy to derive trajectories directly from the joint distribution, effectively grounding robotic control in the generative dynamics of the video backbone.

\subsection{Preliminaries}

\textbf{Flow matching.}
Flow Matching (FM) aims to regress a time-dependent velocity field $v_\theta(x, \tau)$ that transports samples along a probability path between a noise distribution $p_1 = \N(0, I)$ and the data distribution $p_0$~\citep{lipman2022flow}. Specifically, consider a conditional probability path $p_{\tau}(x|x_0)$ constructed via an optimal transport displacement map. The interpolation path is defined as:
\begin{equation}
x_{\tau} = (1-\tau) \cdot x_0 + \tau \cdot z, \quad \tau \in [0, 1],
\label{eq:interpolation}
\end{equation}
where $x_0 \sim p_{\text{data}}$ and $z \sim \N(0, I)$. Under this formulation, $\tau=0$ corresponds to the clean data point $x_0$, while $\tau=1$ denotes pure Gaussian noise $z$. The target velocity (ground truth flow) that generates this linear interpolation is the time derivative:
\begin{equation}
v^*(x_{\tau}, \tau) = \frac{dx_{\tau}}{d\tau} = z - x_0.
\label{eq:target_velocity}
\end{equation}
The training objective of flow matching is to minimize the expected $L_2$ distance between the predicted velocity field $v_\theta$ and the target velocity:
\begin{equation}
\Loss_{\text{FM}} = \E_{x_0, z, \tau} \left[ \left| v_\theta(x_{\tau}, \tau) - (z - x_0) \right|^2 \right],
\label{eq:fm_loss}
\end{equation}
where $\tau$ is sampled uniformly from $\U[0, 1]$.
During inference, sampling is performed by solving the Ordinary Differential Equation (ODE) associated with the learned velocity field $v_\theta$. This process involves integrating $v_\theta$ starting from the noise distribution at $\tau=1$ toward the data distribution at $\tau=0$:
\begin{equation}
\frac{dx}{d\tau} = v_\theta(x, \tau), \quad x_1 \sim \N(0, I).
\end{equation}
We employ a first-order Euler discretization to perform the numerical integration. Given a total of $N$ sampling steps and a constant step size $\Delta \tau = 1/N$, the iterative update rule is formulated as:
\begin{equation}
x_{\tau-\Delta \tau} = x_{\tau} - \Delta \tau \cdot v_\theta(x_{\tau}, \tau).
\label{eq:euler_sampling}
\end{equation}


\textbf{Problem statement.}
Instead of current VLA policies that map directly from observations to actions as $\pi_{\theta} (\mathbf{a}_t \mid \mathbf{o}_{t}, l)$ ($l$ is the language goal), DiT4DiT follows a paradigm of \textit{predicting video dynamics-inverse dynamics}. Specifically, the task is to sample video dynamics from inference of video DiT, and predict the actions by reversing the video dynamics sampled. We formulate the process as:
\begin{align}
 \mathbf{o}_{t+1} & \sim p_{v}(\cdot \mid \mathbf{o}_t, l),   \\
 \mathbf{a}_{t} \sim  p_{a}  (\cdot \mid \mathbf{o}_t, \mathcal{H}(\mathbf{o}_{t+1}^{\tau_v}) & ), \text{where } \mathbf{o}_{t+1}^{\tau_v} \xrightarrow{\tau_v \to 0} \mathbf{o}_{t+1}
\end{align}
where $p_v$ and $p_a$ denote the probability distributions of video generation and action generation, respectively. $\mathbf{o}_{t+1}^{\tau_v}$ means the intermediate state of the future frame at flow step $\tau_v$, reflecting its degree of generation, and $\mathcal{H}$ means the process of extracting hidden states from the generation of $\mathbf{o}_{t+1}^{\tau_v}$. 
The training task is to model the joint probability distribution $p_{va}$ of $p_v$ and $p_a$ like:
\begin{equation}
    \mathbf{o}_{t+1}, \mathbf{a}_{t} \sim p_{va}(\cdot \mid \mathbf{o}_t, l).
\end{equation}

\subsection{Dual-DiT Architecture}

Let $\mathbf{o}_t \in \R^{T_{cond} \times 3 \times H \times W}$ denote the observation frames (conditional input, $T_{cond}$ denote the number of condition frames) and $\mathbf{o}_{t+1} \in \R^{T_v \times 3 \times H \times W}$ denote the ground truth future frames, where $T_v$ represents the horizon of future frames.

\textbf{Video DiT.}
We use the Cosmos-Predict2.5-2B~\citep{cosmos25} as the initialization of our video backbone.
 This backbone consists of two primary components: a causal video VAE and a video diffusion transformer. The spatio-temporal VAE serves as the initial compression stage, mapping high-dimensional pixel-space observations $\mathbf{o}_t, \mathbf{o}_{t+1}$ into a compact latent space via significant spatial and temporal downsampling, denoted as $\mathbf{z}_{t}^{0}, \mathbf{z}_{t+1}^0$. 
 The normalized latents $\mathbf{z}_t^0$ are then processed by the DiT, which utilizes a flow-prediction parameterization and is conditioned on language instructions via multi-layer embeddings from Cosmos-Reason1~\citep{azzolini2025cosmosreason1}. Crucially, rather than utilizing the final denoised video output, we repurpose the DiT~\citep{peebles2023scalable} as a feature extractor: a forward hook mechanism intercepts intermediate hidden activations in flow timestep $\tau_f$—either from a specific deep transformer block or averaged across all layers—converting the generative process into rich visual tokens for downstream tasks. This process is formulated as:

\begin{equation}
\mathbf{h}_t^{\tau_f} = \mathcal{H} \big[ v_{\theta}^{\text{video}} \big] \big( \mathbf{z}_{t+1}^{\tau_f}, \tau_f \mid \mathbf{z}_t^0, l \big), \quad \text{where } \mathbf{z}_{t+1}^{\tau_f} \xrightarrow{\tau_f \to 0} \mathbf{z}_{t+1}^{0} 
\label{eqn:feature}
\end{equation}

where $\mathcal{H}\big[\cdot\big]$ denotes the hook operator that extracts the internal hidden states during the forward pass of the velocity network $v_{\theta}^{\text{video}}$, and $\mathbf{z}_{t+1}^{\tau_f} \xrightarrow{\tau_f \to 0} \mathbf{z}_{t+1}^{0}$ indicates the probability flow toward the clean future latent.

\textbf{Action DiT.}
To decode these visual representations into continuous robot control commands, we employ a dedicated action diffusion transformer adapted from the GR00T-N1~\citep{bjorck2025gr00t}. This component operates as a separate flow-matching model composed of a stack of transformer blocks, each utilizing Adaptive Layer Normalization (AdaLN)~\citep{peebles2023scalable} to inject diffusion timestep information and cross-attention layers to attend to the visual features $\mathbf{h}_t^{\tau_f}$ extracted by the video backbone. The input sequence to this DiT is a concatenation of proprioceptive state embeddings, encoded noisy action trajectories, and a set of learnable ``future tokens" that serve as compressed queries for the motion planning task. Through the cross-attention mechanism, the action head fuses the spatiotemporal visual context with the robot's state, refining the noisy inputs into a coherent trajectory. The network terminates with a linear projection that predicts the velocity vector field of the action sequence, allowing the final trajectory to be synthesized via iterative numerical integration during inference.

\subsection{Joint Training of Video and Action}

To operationalize the simultaneous modeling of latent representations for both video and action, we propose a Dual Flow-Matching mechanism. This approach unifies the generative video prediction and the inverse dynamics of action inference into a single learning paradigm, optimizing both DiTs through a joint objective.

\begin{figure}
    \centering
    \includegraphics[width=0.92\linewidth]{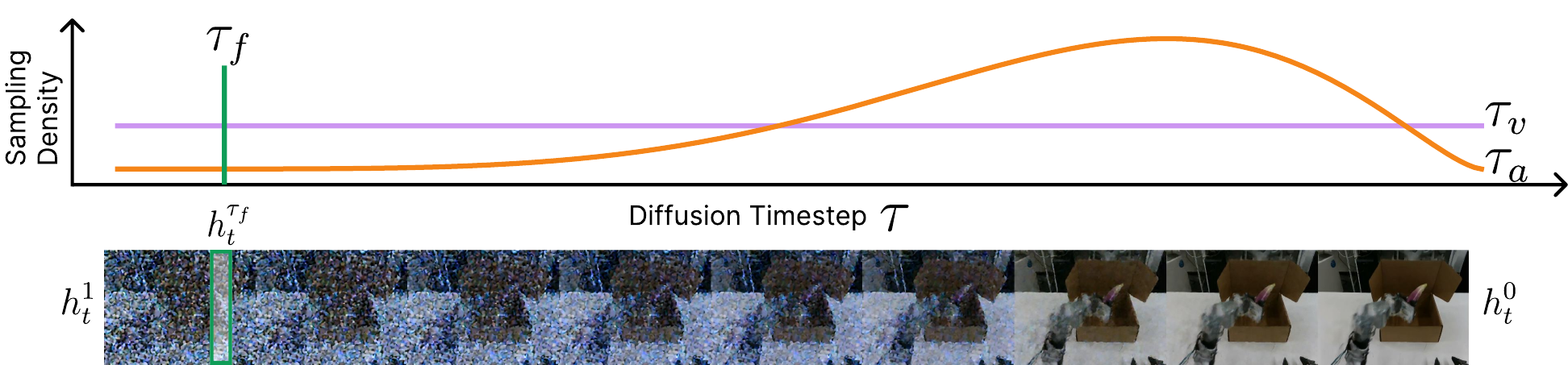}
\caption{\textbf{Asymmetric tri-timestep design.} We decouple the diffusion timesteps to optimize joint video-action generation. The video module uses uniform sampling ($\tau_v$) to capture the full denoising trajectory, while the action module uses Beta sampling ($\tau_a$) to focus on critical control phases. Meanwhile, stable visual conditions are extracted at a fixed deterministic timestep ($\tau_f$) from the evolving hidden states ($h_t^1 \rightarrow h_t^0$).}
\label{fig:tri_timestep}
\vspace{-15pt}
\end{figure}

\textbf{Tri-timestep scheme.}
A core challenge in this joint optimization is balancing the divergent requirements of generative modeling and feature extraction. To address this and achieve the simultaneous modeling of latent representations for both video and action, we adopt an asymmetric \textit{tri-timestep} scheme that decouples the diffusion process of the visual backbone from that of the action module, as shown in Fig.~\ref{fig:tri_timestep}.

For the video generation module, we follow the standard diffusion training~\citep{nvidia2025cosmosworldfoundationmodel,cosmos25} paradigm. At each training step, the prediction timestep $\tau_v$ is randomly sampled from a uniform distribution, $\tau_v \sim \mathcal{U}[0, 1]$. This exposes the model to all noise levels, forcing it to learn the full denoising trajectory required to synthesize future frames.

Conversely, the feature extraction process requires deterministic and consistent representations across iterations to ensure the downstream action module receives a stable input signal. Therefore, when extracting the intermediate representation $\mathbf{h}_t^{\tau_f}$, we forward the context frames through the denoising backbone at a \emph{fixed} timestep, denoted as $\tau_f$. This fixed timestep acts as a conditioning signal, selecting a specific ``operating point'' of the backbone: while early diffusion stages emphasize global structure, later stages attend to fine-grained details. By fixing this value, we stabilize the latent representations, yielding features that are consistently informative for downstream action prediction during both training and inference.

Finally, the action DiT relies on a third, independent timestep, $\tau_a$. Unlike the video generation module which employs uniform sampling, $\tau_a$ is drawn from a Beta distribution during training ($\tau_a = 1 - \sigma$, where $\sigma \sim \text{Beta}(\alpha, \beta)$). This biased continuous-time sampling strategy allocates more training capacity to the most critical stages of the flow trajectory. This complete decoupling allows the action decoder to independently learn the optimal inverse dynamics—mapping pure noise to precise actions—while remaining continuously conditioned on the stable visual features provided by the fixed feature-extraction timestep $\tau_f$.

\textbf{Training.}
The video and action DiTs are jointly fine-tuned to achieve simultaneous modeling of latent representations for both video and action, as detailed in Algorithm~\ref{alg:training}. During training, the text encoder and visual VAE are frozen, restricting the parameter updates entirely to the DiT modules to adapt them to the target domain.
Building upon this dual-timestep design, the overall training objective is formulated as a joint flow-matching loss:

\begin{multline}
    \Loss^{\text{total}}_t = \underbrace{\E_{\tau_a, \epsilon} \left[ \left\| \vaction\left(\mathbf{a}_t^{\tau_a}, \tau_a \mid \mathbf{h}_t^{\tau_f}, s\right) - (\epsilon - \mathbf{a}_t^0) \right\|^2 \right]}_{\text{Action Flow Matching Loss}} \\
    + \lambda \underbrace{\E_{\tau_v, z} \left[ \left\| \vvideo\left(\mathbf{z}_{t+1}^{\tau_v}, \tau_v \mid \mathbf{z}_{t}^{0}, l \right) - (z - \mathbf{z}_{t+1}^{0}) \right\|^2 \right]}_{\text{Video Flow Matching Loss}}
    \label{eq:total_loss_expanded}
\end{multline}
where $\lambda$ is a scalar coefficient that balances the two learning signals. For the video flow-matching objective, the video DiT is trained to predict the velocity via $v_\theta^{\text{video}}$ that transports the current observation $\mathbf{z}_{t}^{0}$ and language goal $l$ toward the future latent state.
For the action flow-matching objective, the action DiT learns to map the noisy action to the target action velocity $\epsilon - \mathbf{a}_t^0$. Crucially, this action prediction is conditioned on the robot's proprioceptive state $s$ and the hidden features $\mathbf{h}_t^{\tau_f}$ extracted from the video backbone in timestep $\tau_f$ as shown in Eqn.~\ref{eqn:feature}. By jointly minimizing these objectives, the framework ensures that the generative dynamics of the visual world inherently scaffold the execution of complex robotic actions.

\begin{algorithm}[]
\caption{Joint Training of Video and Action DiT}
\label{alg:training}
\begin{algorithmic}[1]
\Require Observation $\mathbf{o}_{t}$, future frame $\mathbf{o}_{t+1}$, action $\azero$, state $s$, language goal $l$, action mask $M$
\Ensure Updated parameters $\theta$ (Video DiT), $\phi$ (Action DiT)

\State \textcolor{orange}{\texttt{// ===== Video DiT Forward =====}}
\State $\mathbf{z}_{t}^0 \gets \text{VAE}_{\text{enc}}(\mathbf{o}_{t})$ \Comment{Encode observation}
\State $\mathbf{z}_{t+1}^{0} \gets \text{VAE}_{\text{enc}}(\mathbf{o}_{t+1})$ \Comment{Encode future frames}

\State $\tau_v \sim \U[0, 1]$ \Comment{Sample video timestep}
\State $z \sim \N(0, I)$ \Comment{Sample video noise}
\State $\mathbf{z}_{t+1}^{\tau_v} \gets (1-\tau_v) \cdot \mathbf{z}_{t+1}^{0} + \tau_v \cdot z$ \Comment{Noisy future latent}

\State $\hat{v}_{\text{video}} \gets \vvideo(\mathbf{z}_{t+1}^{\tau_v}, \tau_v \mid \mathbf{z}_{t}^{0}, l)$ \Comment{Predict velocity}
\State $v^*_{\text{video}} \gets z - \mathbf{z}_{t+1}^{0}$ \Comment{Target velocity}
\State $\Loss_{\text{video}} \gets \| \hat{v}_{\text{video}} - v^*_{\text{video}} \|^2$ \Comment{Video loss}

\State \textcolor{orange}{\texttt{// ===== Extract Hidden States =====}}
\State $\tau_f \sim \mathcal{U}\{0/T, 1/T, \dots, T/T\}$ \Comment{Sample hidden extracting timestep}
\State $\hat{\mathbf{z}}_{t+1} \sim \N(0, I) $ \Comment{Sample future noise}
\State $\mathbf{h}_t^{\tau_f} \gets \mathcal{H} (\theta, \hat{\mathbf{z}}_{t+1}, \tau_f, \mathbf{z}_t^0, l)$  \Comment{Extract hidden states}

\State \textcolor{orange}{\texttt{// ===== Action DiT Forward =====}}
\State $\sigma \sim \text{Beta}(\alpha, \beta)$; $\tau_a \gets 1-\sigma$ \Comment{Sample action timestep}
\State $\epsilon \sim \N(0, I)$ \Comment{Sample action noise}
\State $\mathbf{a}_{t}^{\tau_a} \gets (1-\tau_a) \cdot \mathbf{a}_t^{0} + \tau_a \cdot \epsilon$ \Comment{Noisy action}

\State $\hat{v}_{\text{action}} \gets \vaction(\mathbf{a}_t^{\tau_a}, \tau_a \mid \mathbf{h}_t^{\tau_f}, s)$ \Comment{Predict velocity}
\State $v^*_{\text{action}} \gets \epsilon-\mathbf{a}_t^{0}$ \Comment{Target velocity}
\State $\Loss_{\text{action}} \gets \| (\hat{v}_{\text{action}} - v^*_{\text{action}}) \odot M \|^2 / \|M\|_1$ \Comment{Masked action loss}

\State \textcolor{orange}{\texttt{// ===== Backward =====}}
\State $\Loss_{\text{total}} \gets \Loss_{\text{action}} + \lambda \cdot \Loss_{\text{video}}$
\State Update $\theta, \phi$ via $\nabla \Loss_{\text{total}}$

\end{algorithmic}
\end{algorithm}

\subsection{Inference}

During inference, the DiT4DiT framework demonstrates highly flexible generative capabilities, equipped to perform both video generation and action prediction. It executes a decoupled sampling procedure that can synthesize future visual dynamics, infer precise robot control commands, or perform both tasks concurrently, as detailed in Algorithm~\ref{alg:inference}.

\textbf{Video DiT Sampling.} 
When tasked with synthesizing future visual dynamics, the framework activates the video generation pathway. The current observation $\mathbf{o}_t$ is compressed into a latent representation $\mathbf{z}_t^0$ via the frozen VAE encoder. Starting from a standard Gaussian noise distribution $\hat{\mathbf{z}}_{t+1} \sim \mathcal{N}(0, I)$, the video model iteratively updates the latent over $N_v$ discrete steps. At each flow step $\tau_v$, the network predicts the velocity field $\hat{v}$ conditioned on the initial observation $\mathbf{z}_t^0$ and the language goal $l$. The latent is updated using the Euler step rule until it reaches the clean future state, which is subsequently projected back to pixel space via the VAE decoder to yield the predicted future frame $\hat{\mathbf{o}}_{t+1}$.

\textbf{Action DiT Sampling.} 
Rather than relying on the intermediate states of the full video generation loop, the action conditioning requires only a single, deterministic feature extraction step. We sample a new noise latent and perform a single forward pass through the video backbone evaluated strictly at the fixed feature-extraction timestep $\tau_f$. This step intercepts the intermediate activations via the hook mechanism $\mathcal{H}$, yielding a stable and deterministic hidden representation $\mathbf{h}_t^{\tau_f}$. With the visual context established, the action trajectory is initialized from noise $\hat{\mathbf{a}}_t \sim \mathcal{N}(0, I)$. Over $N_a$ numerical integration steps, the Action DiT predicts the action velocity field conditioned on the extracted generative features $\mathbf{h}_t^{\tau_f}$ and the robot's proprioceptive state $s$. The trajectory is refined iteratively, ultimately yielding the precise predicted action $\hat{\mathbf{a}}_t$.

\begin{algorithm}[]
\caption{DiT4DiT Inference}
\label{alg:inference}
\begin{algorithmic}[1]
\Require Observation $\mathbf{o}_{t}$, state $s$, language goal $l$
\Require $N_v$: video sampling steps, $N_a$: action sampling steps
\Ensure Predicted action $\hat{\mathbf{a}_t}$, predicted future frame $\hat{\mathbf{o}}_{t+1}$

\State \textcolor{orange}{\texttt{// ===== Video DiT Sampling =====}}
\State $\mathbf{z}_t^0 \gets \text{VAE}_{\text{enc}}(\mathbf{o}_t)$
\State $\hat{\mathbf{z}}_{t+1} \sim \N(0, I)$ \Comment{Initialize from noise}
\State $\Delta \tau_v \gets 1 / N_v$

\For{$i = 0, 1, \ldots, N_v - 1$}
    \State $\tau_v \gets 1 - i \cdot \Delta \tau_v$
    \State $\hat{v} \gets \vvideo(\hat{\mathbf{z}}_{t+1}, \tau_v \mid \mathbf{z}_t^{0}, l)$
    \State $\hat{\mathbf{z}}_{t+1} \gets \hat{\mathbf{z}}_{t+1} - \Delta \tau_v \cdot \hat{v}$ \Comment{Euler step backward}
\EndFor

\State $\hat{\mathbf{o}}_{t+1} \gets \text{VAE}_{\text{dec}}(\hat{\mathbf{z}}_{t+1})$

\State \textcolor{orange}{\texttt{// ===== Action DiT Sampling =====}}
\State $\hat{\mathbf{a}}_{t} \sim \N(0, I)$ \Comment{Initialize from noise}
\State $\hat{\mathbf{z}}_{t+1} \sim \N(0, I)$ \Comment{Initialize from noise}
\State  $\mathbf{h}_t^{\tau_f} \gets \mathcal{H} (\theta, \hat{\mathbf{z}}_{t+1}, \tau_f, \mathbf{z}_t^0, l)$ \Comment{Extract hidden states}

\State $\Delta \tau \gets 1 / N_a$

\For{$i = 0, 1, \ldots, N_a - 1$}
    \State $\tau_a \gets 1-i \cdot \Delta \tau_a$
    \State $\hat{v} \gets \vaction(\hat{\mathbf{a}}_{t}, \tau_a \mid \mathbf{h}_t^{\tau_f}, s)$
    \State $\hat{\mathbf{a}}_t \gets \hat{\mathbf{a}}_t - \Delta \tau \cdot \hat{v}$ \Comment{Euler step backward}
\EndFor

\State \Return $\hat{\mathbf{a}}_t, \hat{\mathbf{o}}_{t+1}$

\end{algorithmic}
\end{algorithm}

\section{Experiments}

\begin{figure}[htbp]
    \centering
    \includegraphics[width=1.0\linewidth]{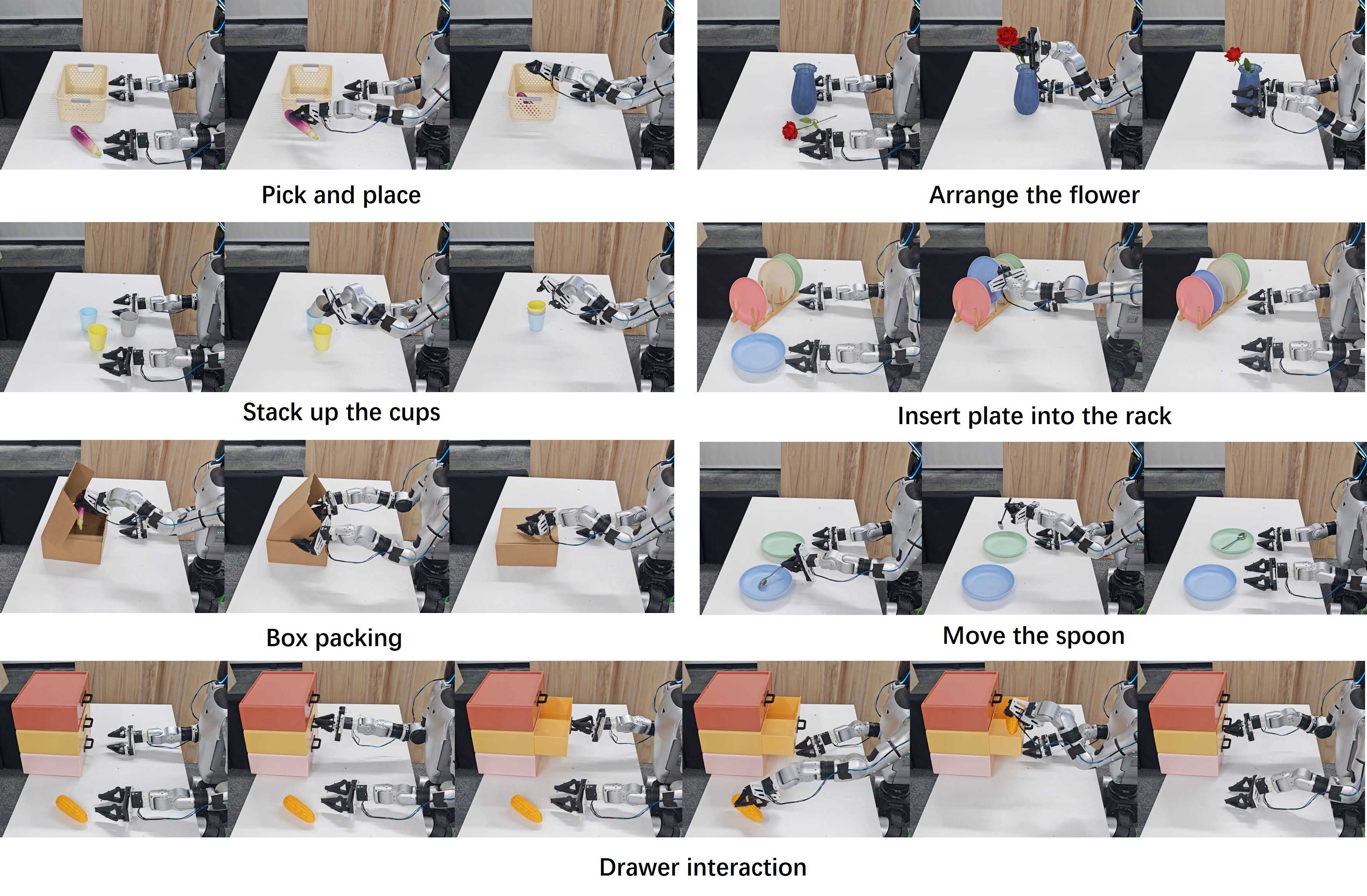}
    \vspace{-15pt}
    \caption{\textbf{Real-world evaluation suite on the Unitree G1 humanoid robot.} The selected tasks evaluate distinct dimensions of robotic proficiency, ranging from high-precision spatial manipulation (e.g., \textit{stack up the cups, insert plate into the rack, arrange the flower}) to complex, extended-horizon execution (e.g., \textit{box packing, drawer interaction)}.}
    \label{fig:real_task}
\vspace{-5pt}
\end{figure}

We evaluate DiT4DiT to address three primary questions: (1) how DiT4DiT compares with state-of-the-art VLM-based policies in both simulation and real-world deployment; (2) whether a VAM with a video-generative backbone offers advantages over a parameter-matched VLM-based VLA baseline; and (3) how well DiT4DiT generalizes under distribution shifts. To answer these questions, we conduct a comprehensive experimental suite covering benchmark comparison, real-world evaluation, zero-shot generalization, and ablation/efficiency analysis across different tasks and robot embodiments.

\subsection{Experiment Setup}
\textbf{LIBERO benchmark.}
 LIBERO benchmark \citep{liu2024libero} focuses on manipulation tasks performed by a Franka Emika Panda manipulator. The evaluation spans four distinct suites—LIBERO-Spatial, LIBERO-Object, LIBERO-Goal, and LIBERO-Long—designed to systematically test a model's proficiency in generalizing to novel spatial configurations, interacting with unseen objects, interpreting language instructions, and executing extended-horizon behaviors, respectively. The standard dataset for each category contains exactly 500 demonstration trajectories, distributed evenly across 10 unique tasks.

\textbf{RoboCasa-GR1 tabletop benchmark.} To further evaluate our approach on a more complex embodiment, we adopt the RoboCasa-GR1 tabletop benchmark \citep{bjorck2025gr00t, nasiriany2024robocasa}. Built upon the RoboCasa simulation framework, this benchmark features the Fourier GR1 humanoid robot equipped with two 7-DoF arms, two 6-DoF Fourier dexterous hands, and a 3-DoF waist, resulting in a 29-dimensional action space. For visual observations, our policy relies exclusively on the robot's egocentric (ego-view) camera. The suite encompasses 24 distinct household manipulation tasks, designed to assess a policy's ability to handle diverse activities ranging from articulated object interaction (e.g., \textit{opening microwaves} or \textit{cabinets}) to complex pick-and-place behaviors with novel objects. The standard dataset provides an extensive collection of teleoperated demonstrations, supplying exactly 1,000 human-collected trajectories for each of the 24 tasks. During evaluation, each task is tested over 50 rollouts with a maximum episode horizon of 720 environment steps. We report the average success rate (\%) across rollouts for each task and the overall average across all 24 tasks. 

\textbf{Real-world G1 tasks.} 
To validate the real-world applicability of our approach, we deploy our policy on a Unitree G1 humanoid robot. The robotic system features a continuous 16-DoF action space, driven by two 7-DoF arms and ALOHA2 grippers, and relies exclusively on the robot's egocentric (ego-view) camera for visual observations. To comprehensively assess the model's robustness across diverse physical interactions and spatial reasoning challenges, we construct a benchmark suite comprising seven distinct household manipulation tasks. As illustrated in Fig.~\ref{fig:real_task}, these include: \textit{pick and place, arrange the flower, stack up the cups, insert plate into the rack, box packaging, move the spoon}, and \textit{drawer interaction}. For each task, we collected a dataset of exactly 200 human demonstration episodes. During the evaluation phase, performance is measured over 20 independent real-world rollouts per task, with the success rate reported as the primary metric.

\textbf{Policy Setup and Baselines.} 
To rigorously evaluate the proposed DiT4DiT framework, we benchmark it against a diverse set of state-of-the-art policies, primarily focusing on the established GR00T series~\citep{bjorck2025gr00t,gr00tn16} and a custom, parameter-matched baseline denoted as Qwen3DiT. This baseline method combines Qwen3-VL~\citep{bai2025qwen3} 2B foundation model with the same action DiT used in DiT4DiT.

For the simulated experiments, we train both DiT4DiT and Qwen3DiT entirely from scratch. This guarantees a strictly fair comparison of their inherent architectural efficiency and learning capabilities, while the remaining external baselines are evaluated using their official open-sourced pre-trained weights.

For the real-world experiments, we employ a two-stage training pipeline. DiT4DiT is first pre-trained on a subset of the simulated GR1 dataset~\citep{bjorck2025gr00t}, comprising 241,450 episodes, to acquire fundamental spatiotemporal priors, followed by fine-tuning on the teleoperated real-world G1 demonstrations. Under this setting, we compare our approach against GR00T-N1.5~\citep{bjorck2025gr00t} and Qwen3DiT. To provide a stringent ablation, Qwen3DiT is subjected to the exact same pre-training and fine-tuning pipeline as DiT4DiT. In contrast, GR00T-N1.5 is initialized from its official pre-trained weights, benefiting from a significantly larger scale of prior data before being fine-tuned on our target real-world tasks. Specifically, our pre-training data volume is  merely $\sim$\textbf{15\%} of the scale of training data leveraged by the official GR00T-N1.5 model.

\subsection{Comparison against State-of-the-art Policies}

\begin{table}[htbp]
\centering
\resizebox{0.9\textwidth}{!}{\begin{tabular}{lccccc}
\toprule
& Spatial SR (\%) & Object SR (\%) & Goal SR (\%) & Long SR (\%) & Average SR (\%)\\
\midrule
Diffusion Policy \citep{chi2023diffusion} & 78.3 & 92.5  & 68.3  & 50.5   & 72.4 \\
Dita \citep{hou2025dita} & 97.4 & 94.8 & 93.2 & 83.6 & 92.3 \\
$\pi_0$ \citep{black2024pi_0} & 96.8 & 98.8 & 95.8 & 85.2 & 94.2 \\
UniVLA \citep{bu2025univla} & 96.5 & 96.8 & 95.6 & 92.0 & 95.2 \\
$\pi_{0.5}$ \citep{intelligence2025pi} & \textbf{98.8} & 98.2 & 98.0 & 92.4 & 96.9 \\
OpenVLA-OFT \citep{kim2025fine} & 97.6 & 98.4 & 97.9 & 94.5 & 97.1  \\
CogVLA \citep{li2025cogvla} & 98.6 & 98.8 & 96.6 & 95.4 & 97.4 \\
GR00T-N1.5~\citep{bjorck2025gr00t} &96.2 &94.0 &96.0 &90.0 &94.1\\
\midrule
Qwen3DiT (\textit{from scratch}) &98.0 &98.8 &96.0 &93.6 &96.6 \\
\rowcolor{yellow!20} DiT4DiT (\textit{from scratch}) &98.4 &\textbf{99.6} &\textbf{98.6} &\textbf{97.6} &\textbf{98.6}  \\

\bottomrule
\end{tabular}}
\caption{\textbf{Success rates (\%) on the four evaluation suites of the LIBERO simulation benchmark.} Bold numbers indicate the highest performance in each category. In this context, \textit{from scratch} implies that the model was trained without using any action data outside of the current benchmark.}
\vspace{-5pt}
\label{tab:libero}
\end{table}

\begin{table}[htbp]
\centering
\resizebox{0.92\textwidth}{!}{
\begin{tabular}{lccc>{\columncolor{yellow!20}}c}
\toprule
\multirow{2}{*}{\textbf{Task}} & \textbf{GR00T-N1.5} & \textbf{GR00T-N1.6} & \textbf{Qwen3DiT} &  \textbf{DiT4DiT} \\
 & \citep{bjorck2025gr00t} & \citep{gr00tn16} &(\textit{from scratch}) &(\textit{from scratch}) \\
\midrule
\textit{BottleToCabinetClose} & 40.0 & 36.0 & \textbf{50.0} & 48.0 \\
\textit{CanToDrawerClose} & 56.0 & 28.0 & 48.0 & \textbf{74.0} \\
\textit{CupToDrawerClose} & 50.0 & 12.0 & 42.0 & \textbf{52.0} \\
\textit{MilkToMicrowaveClose} &\textbf{52.0}  & 20.0 & 38.0 & 50.0 \\
\textit{PotatoToMicrowaveClose} & 22.0 & 28.0 & 18.0 & \textbf{36.0} \\
\textit{WineToCabinetClose} & \textbf{44.0} & 18.0 & 28.0 & 42.0 \\
\addlinespace
\textit{FromCuttingboardToBasket} & 46.0 & 42.0 & 42.0 & \textbf{52.0} \\
\textit{FromCuttingboardToCardboardbox} & 44.0 & 40.0 & 30.0 & \textbf{48.0} \\
\textit{FromCuttingboardToPan} & 58.0 & 62.0 & 50.0 & \textbf{76.0} \\
\textit{FromCuttingboardToPot} & 48.0 & 60.0 & 44.0 & \textbf{62.0} \\
\textit{FromCuttingboardToTieredbasket} & 28.0 & 48.0 & 36.0 & \textbf{50.0} \\
\addlinespace
\textit{FromPlacematToBasket} & 32.0 & 42.0 & 14.0 & \textbf{50.0} \\
\textit{FromPlacematToBowl} & 52.0 & 34.0 & 28.0 & \textbf{56.0} \\
\textit{FromPlacematToPlate} & \textbf{42.0} & \textbf{42.0} & 40.0 & 32.0 \\
\textit{FromPlacematToTieredshelf} & 26.0 & 24.0 & \textbf{30.0} &18.0 \\
\addlinespace
\textit{FromPlateToBowl} & 38.0 & 48.0 & 36.0 & \textbf{56.0} \\
\textit{FromPlateToCardboardbox} & 40.0 & 44.0 & 36.0 & \textbf{58.0} \\
\textit{FromPlateToPan} & 56.0 & 48.0 & 34.0 & \textbf{68.0} \\
\textit{FromPlateToPlate} & 50.0 & \textbf{66.0} & 44.0 & 58.0 \\
\addlinespace
\textit{FromTrayToCardboardbox} & 36.0 & 42.0 & \textbf{48.0} & 38.0 \\
\textit{FromTrayToPlate} & 54.0 & 52.0 & 44.0 & \textbf{56.0} \\
\textit{FromTrayToPot} & 36.0 & \textbf{64.0} & 34.0 & 54.0 \\
\textit{FromTrayToTieredbasket} & 34.0 & 42.0 & 36.0 & \textbf{46.0} \\
\textit{FromTrayToTieredshelf} & 22.0 & \textbf{38.0} & 18.0 & \textbf{38.0} \\
\midrule
\textbf{Average} & 41.8 & 40.8 & 36.2 & \textbf{50.8} \\
\bottomrule
\vspace{-10pt}
\end{tabular}
}
\caption{\textbf{RoboCasa-GR1 tabletop tasks evaluation results} (success rate (\%)). Bold numbers indicate the highest performance in each category. \textit{from scratch} implies that the model was trained without using any action data outside of the current benchmark. While the GR00T models are fine-tuned from their pre-trained weights, they are trained for the exact same number of steps as Qwen3DiT and DiT4DiT, which are trained from scratch.}
\vspace{-5pt}
\label{tab:robocasa}
\end{table}

\textbf{LIBERO benchmark results.} 
Table~\ref{tab:libero} presents the quantitative evaluation of our method alongside state-of-the-art baselines on the LIBERO~\citep{liu2024libero} simulation benchmark. Overall, our proposed DiT4DiT trained from scratch achieves a new state-of-the-art average success rate of 98.6\%, outperforming previous VLA models pre-trained in large-scale action datasets. 

When analyzing the distinct suites, DiT4DiT demonstrates exceptional generalization capabilities across novel objects and language instructions, achieving the highest success rates in the LIBERO-Object (99.6\%) and LIBERO-Goal (98.6\%) suites. Furthermore, our method exhibits a particularly striking advantage on the LIBERO-Long suite, which evaluates extended-horizon behaviors. DiT4DiT attains a 97.6\% success rate on this challenging suite, significantly surpassing the next best method. This robust long-horizon performance strongly validates our design choice: by explicitly modeling the spatiotemporal dynamics through the video DiT backbone, our policy gains a deeper understanding of physical state transitions, which is crucial for executing complex, multi-stage manipulation tasks. 
Finally, compared to our direct baseline, Qwen3DiT (96.6\% average), DiT4DiT yields consistent improvements across all four categories, confirming that our decoupled generative inverse dynamics and feature extraction mechanism successfully translate rich video priors into precise robotic control.

\textbf{RoboCasa-GR1 tabletop benchmark results.} 
We evaluate on 24 challenging manipulation tasks from the RoboCasa-GR1 tabletop suite~\citep{nasiriany2024robocasa}. As summarized in Table~\ref{tab:robocasa}, DiT4DiT achieves a new state-of-the-art average success rate of 50.8\%. This significantly outperforms established, highly optimized policies, exceeding GR00T-N1.5 and GR00T-N1.6 by a substantial margin of 9.0 and 10.0 absolute percentage points, respectively. 

Crucially, we compare DiT4DiT against our parameter-matched direct baseline, Qwen3DiT (36.2\%). 
DiT4DiT demonstrates a remarkable 14.6\% absolute improvement over Qwen3DiT. This substantial leap strongly validates our core hypothesis: substituting static image-text priors with the implicit spatiotemporal dynamics of a generative video model provides a superior conditioning signal for learning complex inverse dynamics.

Examining individual tasks, DiT4DiT exhibits dominant performance, achieving the highest success rate in 16 out of the 24 evaluated tasks. The performance gains are particularly pronounced in tasks demanding precise spatial coordination and complex physical interaction. For instance, on \textit{CanToDrawerClose} (74.0\% vs. 56.0\%), \textit{FromCuttingboardToPan} (76.0\% vs. 62.0\%), and \textit{FromPlateToPan} (68.0\% vs. 56.0\%), our method eclipses the strongest baselines by margins exceeding 12\%.

\begin{figure}[]
    \centering
    \includegraphics[width=0.9\linewidth]{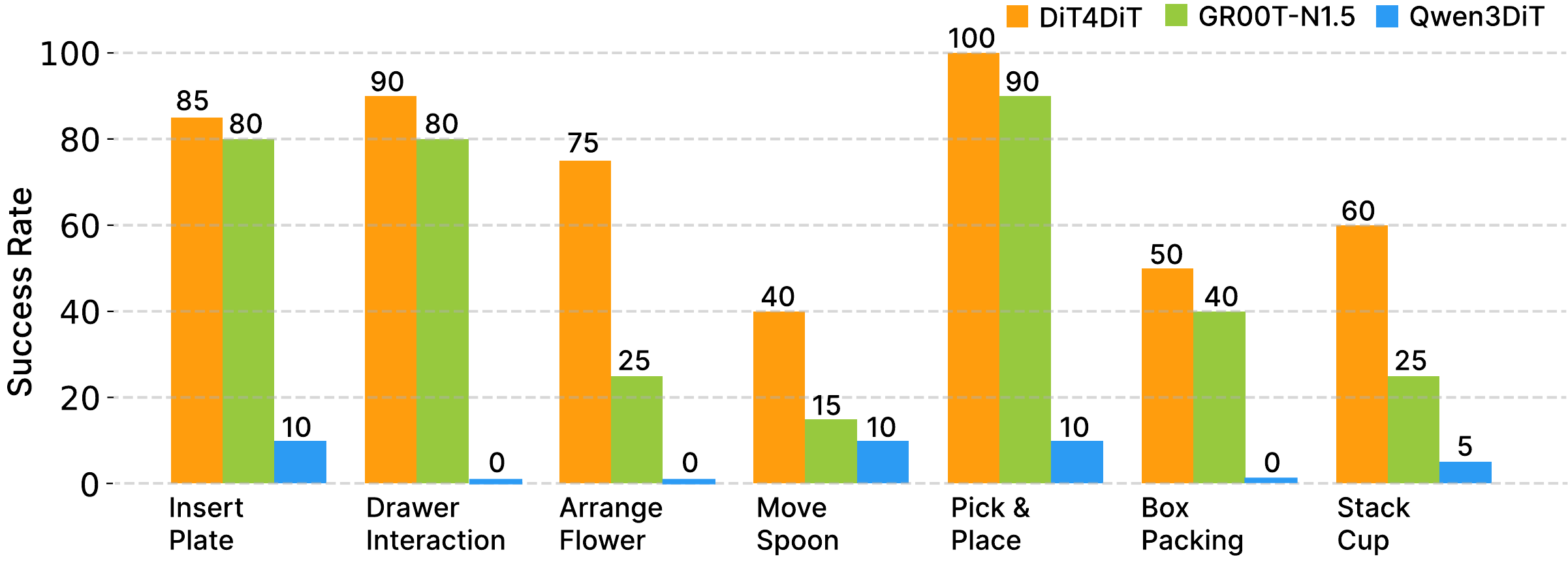}
    \caption{\textbf{Real-world evaluation results on the Unitree G1 robot}. Success rates are reported across seven diverse household tasks. DiT4DiT comprehensively outperforms both the pre-trained GR00T-N1.5~\citep{bjorck2025gr00t} and the parameter-matched Qwen3DiT baseline, highlighting the efficiency and efficacy of our framework.}
    \label{fig:real_task_res}
    \vspace{-20pt}
\end{figure}

\textbf{Real-world G1 task results.}
We report the quantitative success rates across the seven real-world tasks in Fig.~\ref{fig:real_task_res}. Overall, DiT4DiT demonstrates dominant and robust performance, comprehensively outperforming both the pre-trained GR00T-N1.5~\citep{bjorck2025gr00t} and our baseline, Qwen3DiT.

The Qwen3DiT baseline suffers a near-total collapse in the real world. It fails to exceed a 10\% success rate on any task and scores 0\% on Drawer Interaction, Arrange Flower, and Box Packing. This failure underscores the limitation of static image-text priors: without massive real-world trajectory data, VLMs struggle to ground visual semantics into continuous 3D physical actions. In stark contrast, DiT4DiT successfully abstracts robust, physics-aware representations from the simulated pre-training phase, enabling highly effective transfer to the physical robot.

When compared against GR00T-N1.5 (which benefits from a significantly larger scale of pre-training data), DiT4DiT still maintains a consistent lead, with particularly massive margins in tasks demanding high-precision spatial coordination. For example, in the \textit{Arrange Flower} task, which requires delicate alignment to insert a thin stem into a vase, DiT4DiT achieves a 75\% success rate, outperforming GR00T-N1.5 (25\%). We observe similarly compelling gaps in \textit{Stack Cup} (60\% vs. 25\%) and \textit{Move Spoon} (40\% vs. 15\%). We hypothesize that the generative video backbone, by learning to predict future visual representations, inherently preserves more fine-grained visual details than standard VLA policy.

Furthermore, DiT4DiT excels in extended-horizon and multi-stage reasoning tasks. On \textit{Drawer Interaction} and \textit{Box Packing}, which require the robot to sequence multiple distinct sub-goals (e.g., opening a flap, inserting an object, and retreating), DiT4DiT achieves 90\% and 50\% success rates, respectively. By intercepting intermediate denoising features that naturally encode future dynamic transitions, our tri-timestep mechanism equips the action policy with superior temporal consistency, ensuring stable execution over long physical horizons.

\subsection{Generalization Capability}

\begin{figure}[htbp]
    \centering
    \includegraphics[width=0.9\linewidth]{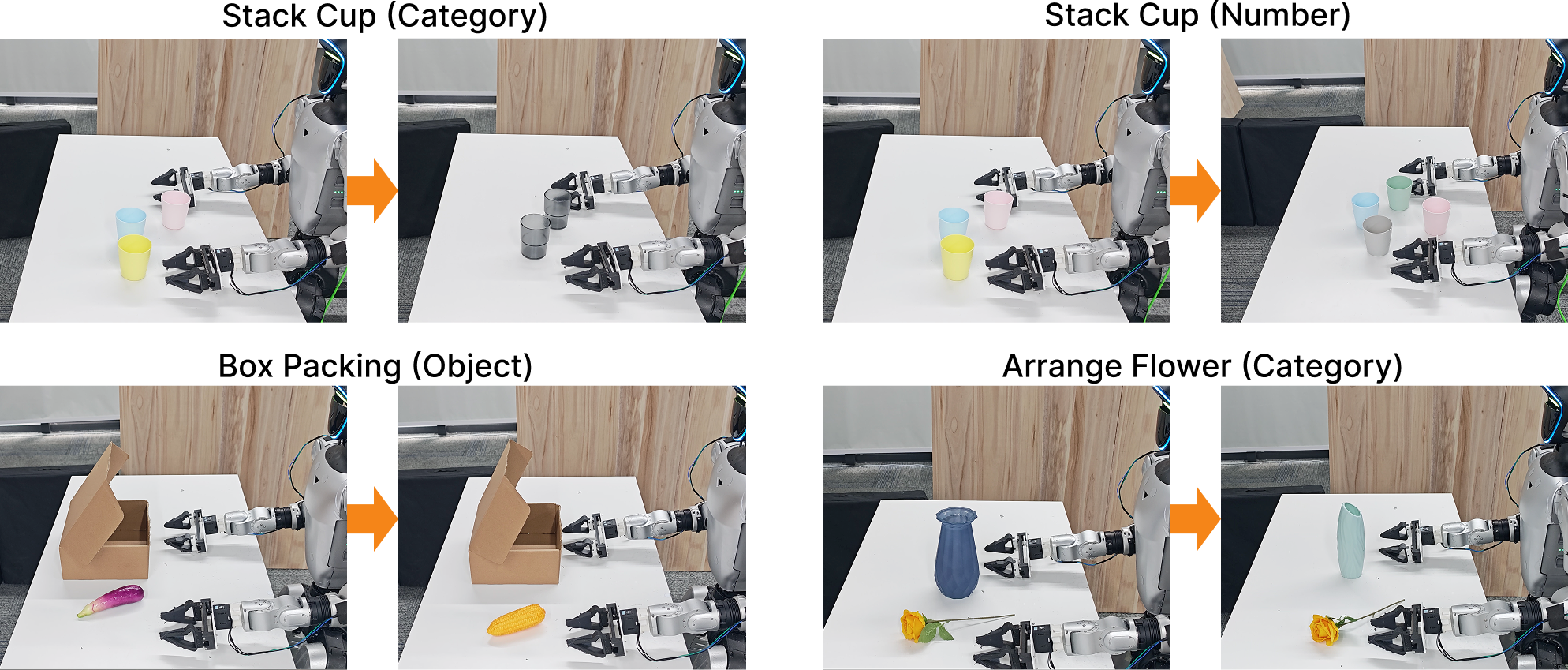}
    \caption{\textbf{Qualitative rollouts of real-world generalization tasks.} We evaluate the policy's zero-shot robustness against severe out-of-distribution physical variations. These include semantic and geometric shifts in \textit{Category} (unseen cups and vases), complete object substitution in \textit{Object} (packing corn instead of an eggplant), and scene clutter in \textit{Number} (stacking four cups instead of three). }
    \label{fig:real_task_gene_exp}
    \vspace{-5pt}
\end{figure}

We evaluate DiT4DiT in both simulation and real-world physical deployments to demonstrate its robust generalization.

\begin{figure}[htbp]
    \centering
    \includegraphics[width=1.0\linewidth]{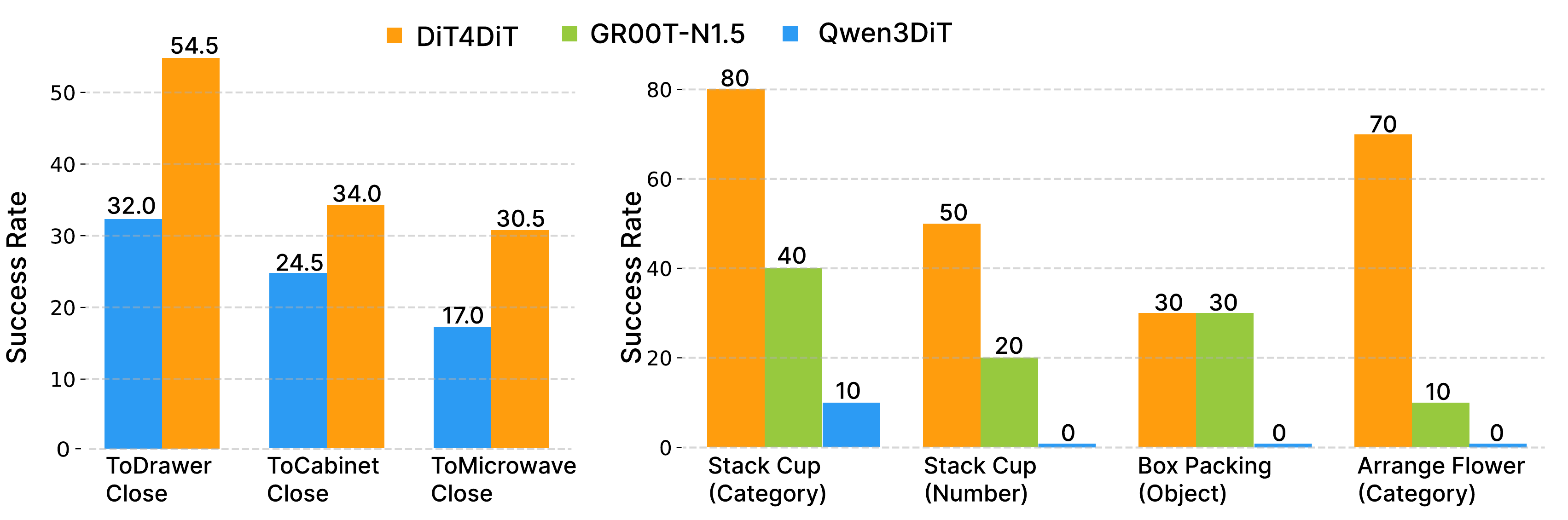}
    \caption{\textbf{Quantitative results of zero-shot generalization.} (\textbf{Left}) Success rates in the simulated RoboCasa-GR1 environment when evaluated on entirely unseen objects. (\textbf{Right}) Success rates on the real-world Unitree G1 robot across four challenging out-of-distribution scenarios, testing category variations, novel object substitutions, and distractor quantities. DiT4DiT demonstrates superior robustness and physical abstraction compared to both the parameter-matched Qwen3DiT baseline and the pre-trained GR00T-N1.5~\citep{bjorck2025gr00t}.}
    \label{fig:task_gene_res}
    \vspace{-10pt}
\end{figure}

In the simulator, we designed a targeted object-substitution experiment within the RoboCasa~\citep{nasiriany2024robocasa} environment. Specifically, we restricted the training distribution to three tasks involving only a single object category: \textit{BottleToDrawerClose, BottleToCabinetClose,} and \textit{BottleToMicrowaveClose}. During evaluation, we completely removed the bottle and tested the policies zero-shot on four unseen objects: \textit{Can, Cup, Milk,} and \textit{Wine}.
DiT4DiT exhibits a remarkably stronger ability to generalize to novel objects compared to the parameter-matched Qwen3DiT baseline, as shown in the left panel of Fig.~\ref{fig:task_gene_res}. Across all three tasks, our approach maintains robust performance. Most notably, in the \textit{ToDrawerClose} tasks, DiT4DiT achieves an impressive 54.5\% success rate with the unseen objects, exceeding Qwen3DiT by a massive 22.5\%. We observe similarly substantial margins in \textit{ToCabinetClose} (34.0\% vs. 24.5\%) and \textit{ToMicrowaveClose} (30.5\% vs. 17.0\%).

To further validate these findings under complex physical dynamics, we introduced four challenging zero-shot generalization scenarios on the real-world Unitree G1 robot (visualized in Fig.~\ref{fig:real_task_gene_exp}). These tasks evaluate three distinct dimensions of generalization:

\begin{itemize}
    \item \textbf{Category generalization}: In \textit{Stack Cup (Category)} and \textit{Arrange Flower (Category)}, we drastically alter the material, shape, and visual appearance of the interactive objects (e.g., swapping standard plastic cups for metallic/glass variants, and changing both the vase and the flower).

    \item \textbf{Object substitution}: In \textit{Box Packing (Object)}, the target item is completely replaced with a novel, out-of-distribution object (e.g., swapping an eggplant for corn).

    \item \textbf{Quantity variation}: In \textit{Stack Cup (Number)}, we test whether the policy can handle a different number of objects than seen during training, evaluating its resistance to distractors and novel scene clutter.
\end{itemize}

As shown in the right panel of Fig.~\ref{fig:task_gene_res}, DiT4DiT demonstrates dominant zero-shot transfer capabilities in the physical world. The performance of the parameter-matched Qwen3DiT baseline completely collapses when faced with real-world visual shifts, scoring 0\% on three out of the four tasks. In contrast, DiT4DiT successfully abstracts the underlying physical interactions, such as the spatial constraints of inserting a stem into a vase or aligning cups. Notably, on the \textit{Arrange Flower (Category)} task, DiT4DiT achieves a 70\% success rate, outperforming the Qwen3DiT (0\%) the pre-trained GR00T-N1.5 (10\%) by large margins. Even when modifying the quantity of objects in \textit{Stack Cup (Number)}, DiT4DiT maintains a 50\% success rate, proving that the generative video representations provide a robust, physics-aware understanding of the scene that is fundamentally invariant to surface-level visual changes or distractor counts.

\subsection{Ablations}

\begin{figure}[htbp]
    \centering
    \includegraphics[width=1.0\textwidth]{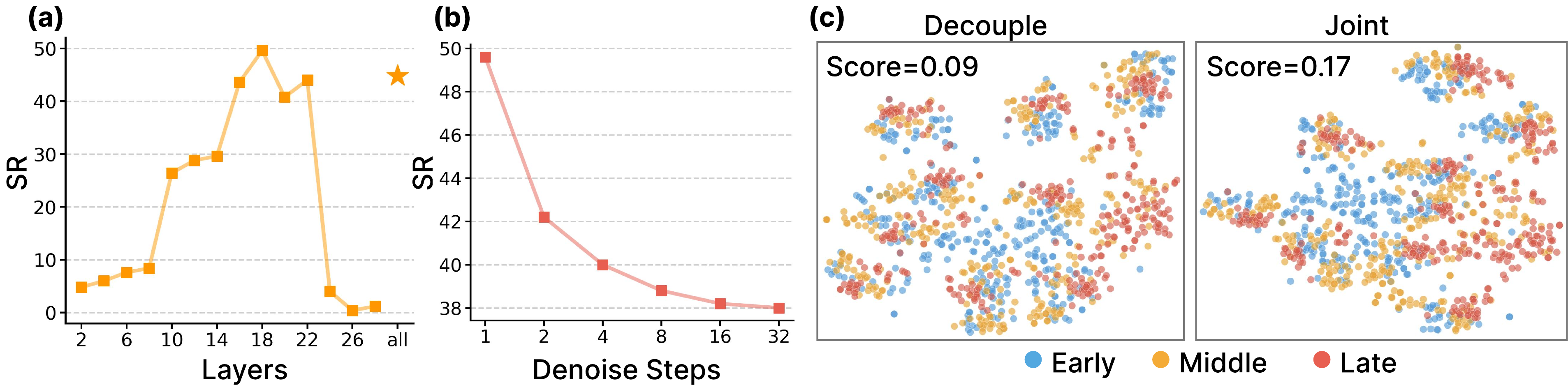} 
    \caption{\textbf{Ablation studies on the DiT4DiT architecture.} (a) \textbf{Feature extraction layer:} Success rate across different hidden layers of the video backbone, with performance peaking at layer 18. (b) \textbf{Denoising steps:} Impact of the number of iterative denoising steps used for action conditioning. A single forward step yields the highest success rate, preventing over-commitment to pixel-level reconstruction. (c) \textbf{Representation learning:} t-SNE visualization of latent features colored by the execution phase (Early, Middle, Late). 
    Supported by a roughly twofold increase in the silhouette score~\citep{rousseeuw1987silhouettes}, our joint training objective successfully induces smooth temporal flows within each task cluster, transitioning fluidly from the Early (blue) through the Middle (yellow) to the Late (red) phases. }
    \label{fig:ablation}
    \vspace{-5pt}
\end{figure}

\textbf{Choice of feature extraction layer.} 
To determine the optimal visual representation for action conditioning, we evaluate the impact of extracting hidden states from different transformer blocks within the Video DiT. We conduct our experiments on five tasks selected from the RoboCasa-GR1 benchmark (\textit{CanToDrawerClose, FromCuttingboardToBasket, FromPlacematToBowl, FromPlateToCardboardbox, FromTrayToPot}).
As illustrated in Figure~\ref{fig:ablation}(a), 
the choice of extraction layer significantly influences the downstream success rate. Features from early layers (e.g., layers 2--8) yield poor performance, likely because they primarily encode low-level visual textures lacking actionable semantics. Performance steadily improves and peaks at layer 18, suggesting that middle-to-deep blocks strike the optimal balance, capturing the rich spatiotemporal physics and high-level scene understanding required for control. Interestingly, extracting from the final layers (layers 24--28) leads to a drastic performance collapse. The result suggests these terminal layers become overly specialized for the immediate video denoising and pixel-level reconstruction objective, thereby discarding abstract, control-relevant representations. Finally, while averaging features across all layers (``all'') yields highly competitive results, it falls slightly short of the single best layer. Consequently, we select layer 18 as the default extraction point for our framework.

\textbf{Denoise steps for hidden features.}
We investigate the impact of the number of iterative denoising steps used to extract the visual hidden features for the action policy. As shown in Figure~\ref{fig:ablation}(b), we evaluate the average success rate across the five selected RoboCasa tasks while varying the denoising steps from 1 to 32. Interestingly, a single denoising step yields the highest performance, with the success rate monotonically degrading as the number of steps increases. The reason could be that excessive iterative denoising forces the hidden states to over-commit to the pixel-level details of a specific reconstructed future. So generalized robust action priors start to lose information with more steps. 
This intuition aligns with findings in recent work~\citep{pai2025mimic}, but this phenomenon is significantly more pronounced in our model, manifesting as a strict monotonic decline. We hypothesize that this extreme sensitivity is a direct consequence of our joint training paradigm: because the video and action generation are updated simultaneously, the action loss heavily regularizes the latent space to extract actionable semantics immediately at the first step, making it highly susceptible to the over-commitment of any subsequent denoising iterations.
Crucially, this finding validates that high-frequency control can be achieved through a single forward pass, entirely bypassing the computational bottleneck of multi-step video generation.

\textbf{Joint \textit{v.s.} decouple training.} 
Finally, we analyze the representational benefits of our joint optimization paradigm. Figure~\ref{fig:ablation}(c) provides a t-SNE visualization of the extracted hidden features, color-coded by their temporal phase within an episode (\textit{Early, Middle, Late}). Under a decoupled training scheme, where the video generative model and the action policy are optimized independently, the features form clusters but exhibit fragmented and entangled temporal distributions within those clusters. Quantitatively, this enhanced temporal separation is reflected by a nearly twofold improvement in the silhouette score (increasing from 0.09 to 0.17). 
Visually, the latent features within each cluster exhibit relatively clear boundaries as they progress from the \textit{Early} to \textit{Middle} to \textit{Late} phases. This qualitative evidence strongly corroborates our core hypothesis: joint training forces the visual backbone to embed a continuous, physics-aware temporal progression, directly empowering the action policy to reason about long-horizon execution and state transitions.

\subsection{Efficiency Analysis}

\begin{table}[htbp]
\small
    \centering
    \begin{tabular}{lcc}
        \toprule
         & \textbf{Trainable Param Cnt} & \textbf{Deploy Frequency} \\
        \midrule
        GR00T-N1.5 & 2.7B & 13Hz \\
        Qwen3DiT & 2.3B & 9Hz  \\
        DiT4DiT      & 2.2B & 6Hz  \\
        \bottomrule
    \end{tabular}
    \caption{\textbf{Deployment efficiency comparison}. We report the trainable parameter count and real-world deployment frequency (Hz) for DiT4DiT and the primary baselines. While the video backbone introduces a computational trade-off yielding a 6Hz control rate (tested on a single NVIDIA A100 GPU), DiT4DiT remains the most parameter-efficient model and comfortably supports real-time closed-loop physical execution.}
    \label{tab:model_comparison}
\end{table}

To assess deployment feasibility, we compare trainable parameter count and real-world control frequency against core baselines (Table~\ref{tab:model_comparison}). DiT4DiT has 2.2B trainable parameters, comparable to Qwen3DiT (2.3B) and smaller than GR00T-N1.5 (2.7B), indicating that its gains do not come from larger model size.
DiT4DiT runs at 6Hz on the physical robot, versus 9Hz for Qwen3DiT and 13Hz for GR00T-N1.5. Although slower, this trade-off is expected for a video-generative backbone that extracts temporally grounded features. Unlike the two baselines, DiT4DiT does not train the LLM component during policy learning; therefore, for fixed tasks, the LLM features remain constant and can be pre-extracted and cached to avoid repeated inference, which can further improve the effective deployment frequency.

\section{Conclusion}

We present DiT4DiT, an end-to-end Video-Action Model that unifies a video DiT and an action DiT through a dual flow-matching objective. Instead of depending on fully reconstructed future frames, our method leverages temporally grounded intermediate denoising features to condition action prediction, enabling physics-aware and stable continuous control.
Across both simulation and real-world experiments, DiT4DiT consistently outperforms strong VLA baselines. It achieves state-of-the-art average success rates on LIBERO and RoboCasa-GR1 (98.6\% and 50.8\%), and maintains strong transfer performance on the Unitree G1 humanoid robot under real-world dynamics.
Beyond aggregate performance, DiT4DiT shows improved robustness under challenging distribution shifts, including unseen object categories and object/scene variations. Overall, our results indicate that modeling video dynamics provides a more effective and data-efficient scaling proxy for policy learning than static image-text priors, offering a practical path toward more generalizable embodied agents.

\bibliography{main_conference}

@inproceedings{gr00tn16,
  archivePrefix = {arxiv},
  eprint     = {2503.14734},
  title      = {{GR00T} {N1}: An Open Foundation Model for Generalist Humanoid Robots},
  author     = {NVIDIA and Johan Bjorck and Fernando Castañeda, Nikita Cherniadev and Xingye Da and Runyu Ding and Linxi "Jim" Fan and Yu Fang and Dieter Fox and Fengyuan Hu and Spencer Huang and Joel Jang and Zhenyu Jiang and Jan Kautz and Kaushil Kundalia and Lawrence Lao and Zhiqi Li and Zongyu Lin and Kevin Lin and Guilin Liu and Edith Llontop and Loic Magne and Ajay Mandlekar and Avnish Narayan and Soroush Nasiriany and Scott Reed and You Liang Tan and Guanzhi Wang and Zu Wang and Jing Wang and Qi Wang and Jiannan Xiang and Yuqi Xie and Yinzhen Xu and Zhenjia Xu and Seonghyeon Ye and Zhiding Yu and Ao Zhang and Hao Zhang and Yizhou Zhao and Ruijie Zheng and Yuke Zhu},
  month      = {March},
  year       = {2025},
  booktitle  = {ArXiv Preprint},
}

@article{azzolini2025cosmosreason1,
  title={Cosmos-reason1: From physical common sense to embodied reasoning},
  author={Azzolini, Alisson and Bai, Junjie and Brandon, Hannah and Cao, Jiaxin and Chattopadhyay, Prithvijit and Chen, Huayu and Chu, Jinju and Cui, Yin and Diamond, Jenna and Ding, Yifan and others},
  journal={arXiv preprint arXiv:2503.15558},
  year={2025}
}

@article{cai2025z,
  title={Z-image: An efficient image generation foundation model with single-stream diffusion transformer},
  author={Cai, Huanqia and Cao, Sihan and Du, Ruoyi and Gao, Peng and Hoi, Steven and Hou, Zhaohui and Huang, Shijie and Jiang, Dengyang and Jin, Xin and Li, Liangchen and others},
  journal={arXiv preprint arXiv:2511.22699},
  year={2025}
}

@article{bi2025motus,
  title={Motus: A unified latent action world model},
  author={Bi, Hongzhe and Tan, Hengkai and Xie, Shenghao and Wang, Zeyuan and Huang, Shuhe and Liu, Haitian and Zhao, Ruowen and Feng, Yao and Xiang, Chendong and Rong, Yinze and others},
  journal={arXiv preprint arXiv:2512.13030},
  year={2025}
}

@article{cosmospolicy,
  title={Cosmos policy: Fine-tuning video models for visuomotor control and planning},
  author={Kim, Moo Jin and Gao, Yihuai and Lin, Tsung-Yi and Lin, Yen-Chen and Ge, Yunhao and Lam, Grace and Liang, Percy and Song, Shuran and Liu, Ming-Yu and Finn, Chelsea and others},
  journal={arXiv preprint arXiv:2601.16163},
  year={2026}
}

@article{pai2025mimic,
  title={mimic-video: Video-action models for generalizable robot control beyond vlas},
  author={Pai, Jonas and Achenbach, Liam and Montesinos, Victoriano and Forrai, Benedek and Mees, Oier and Nava, Elvis},
  journal={arXiv preprint arXiv:2512.15692},
  year={2025}
}

@article{du2023learning,
  title={Learning universal policies via text-guided video generation},
  author={Du, Yilun and Yang, Sherry and Dai, Bo and Dai, Hanjun and Nachum, Ofir and Tenenbaum, Josh and Schuurmans, Dale and Abbeel, Pieter},
  journal={Advances in neural information processing systems},
  volume={36},
  pages={9156--9172},
  year={2023}
}

@article{ye2024latent,
  title={Latent action pretraining from videos},
  author={Ye, Seonghyeon and Jang, Joel and Jeon, Byeongguk and Joo, Sejune and Yang, Jianwei and Peng, Baolin and Mandlekar, Ajay and Tan, Reuben and Chao, Yu-Wei and Lin, Bill Yuchen and others},
  journal={arXiv preprint arXiv:2410.11758},
  year={2024}
}

@article{shen2025videovla,
  title={Videovla: Video generators can be generalizable robot manipulators},
  author={Shen, Yichao and Wei, Fangyun and Du, Zhiying and Liang, Yaobo and Lu, Yan and Yang, Jiaolong and Zheng, Nanning and Guo, Baining},
  journal={arXiv preprint arXiv:2512.06963},
  year={2025}
}

@article{cen2025worldvla,
  title={Worldvla: Towards autoregressive action world model},
  author={Cen, Jun and Yu, Chaohui and Yuan, Hangjie and Jiang, Yuming and Huang, Siteng and Guo, Jiayan and Li, Xin and Song, Yibing and Luo, Hao and Wang, Fan and others},
  journal={arXiv preprint arXiv:2506.21539},
  year={2025}
}

@article{rousseeuw1987silhouettes,
  title={Silhouettes: a graphical aid to the interpretation and validation of cluster analysis},
  author={Rousseeuw, Peter J},
  journal={Journal of computational and applied mathematics},
  volume={20},
  pages={53--65},
  year={1987},
  publisher={Elsevier}
}

@article{lingbot,
  title={Causal World Modeling for Robot Control},
  author={Li, Lin and Zhang, Qihang and Luo, Yiming and Yang, Shuai and Wang, Ruilin and Han, Fei and Yu, Mingrui and Gao, Zelin and Xue, Nan and Zhu, Xing and others},
  journal={arXiv preprint arXiv:2601.21998},
  year={2026}
}

@article{intelligence2025pi06,
  title={a VLA That Learns From Experience},
  author={Intelligence, Physical and Amin, Ali and Aniceto, Raichelle and Balakrishna, Ashwin and Black, Kevin and Conley, Ken and Connors, Grace and Darpinian, James and Dhabalia, Karan and DiCarlo, Jared and others},
  journal={arXiv preprint arXiv:2511.14759},
  year={2025}
}

@article{bai2025qwen3,
  title={Qwen3-vl technical report},
  author={Bai, Shuai and Cai, Yuxuan and Chen, Ruizhe and Chen, Keqin and Chen, Xionghui and Cheng, Zesen and Deng, Lianghao and Ding, Wei and Gao, Chang and Ge, Chunjiang and others},
  journal={arXiv preprint arXiv:2511.21631},
  year={2025}
}

@article{cosmos25,
  title={World simulation with video foundation models for physical ai},
  author={Ali, Arslan and Bai, Junjie and Bala, Maciej and Balaji, Yogesh and Blakeman, Aaron and Cai, Tiffany and Cao, Jiaxin and Cao, Tianshi and Cha, Elizabeth and Chao, Yu-Wei and others},
  journal={arXiv preprint arXiv:2511.00062},
  year={2025}
}

@inproceedings{finn2017deep,
  title={Deep visual foresight for planning robot motion},
  author={Finn, Chelsea and Levine, Sergey},
  booktitle={2017 IEEE international conference on robotics and automation (ICRA)},
  pages={2786--2793},
  year={2017},
  organization={IEEE}
}

@article{ebert2018visual,
  title={Visual foresight: Model-based deep reinforcement learning for vision-based robotic control},
  author={Ebert, Frederik and Finn, Chelsea and Dasari, Sudeep and Xie, Annie and Lee, Alex and Levine, Sergey},
  journal={arXiv preprint arXiv:1812.00568},
  year={2018}
}

@inproceedings{yang2023unisim,
  title={Unisim: A neural closed-loop sensor simulator},
  author={Yang, Ze and Chen, Yun and Wang, Jingkang and Manivasagam, Sivabalan and Ma, Wei-Chiu and Yang, Anqi Joyce and Urtasun, Raquel},
  booktitle={Proceedings of the IEEE/CVF Conference on Computer Vision and Pattern Recognition},
  pages={1389--1399},
  year={2023}
}

@article{achiam2023gpt,
  title={Gpt-4 technical report},
  author={Achiam, Josh and Adler, Steven and Agarwal, Sandhini and Ahmad, Lama and Akkaya, Ilge and Aleman, Florencia Leoni and Almeida, Diogo and Altenschmidt, Janko and Altman, Sam and Anadkat, Shyamal and others},
  journal={arXiv preprint arXiv:2303.08774},
  year={2023}
}

@inproceedings{brohan2023can,
  title={Do as i can, not as i say: Grounding language in robotic affordances},
  author={Brohan, Anthony and Chebotar, Yevgen and Finn, Chelsea and Hausman, Karol and Herzog, Alexander and Ho, Daniel and Ibarz, Julian and Irpan, Alex and Jang, Eric and Julian, Ryan and others},
  booktitle={Conference on robot learning},
  pages={287--318},
  year={2023},
  organization={PMLR}
}

@article{brohan2023rt,
  title={Rt-2: Vision-language-action models transfer web knowledge to robotic control},
  author={Brohan, Anthony and Brown, Noah and Carbajal, Justice and Chebotar, Yevgen and Chen, Xi and Choromanski, Krzysztof and Ding, Tianli and Driess, Danny and Dubey, Avinava and Finn, Chelsea and others},
  journal={arXiv preprint arXiv:2307.15818},
  year={2023}
}

@article{kim2024openvla,
  title={OpenVLA: An Open-Source Vision-Language-Action Model},
  author={Kim, Moo Jin and Pertsch, Karl and Karamcheti, Siddharth and Xiao, Ted and Balakrishna, Ashwin and Nair, Suraj and Rafailov, Rafael and Foster, Ethan and Lam, Grace and Sanketi, Pannag and others},
  journal={arXiv preprint arXiv:2406.09246},
  year={2024}
}

@article{chi2023diffusion,
  title={Diffusion policy: Visuomotor policy learning via action diffusion},
  author={Chi, Cheng and Xu, Zhenjia and Feng, Siyuan and Cousineau, Eric and Du, Yilun and Burchfiel, Benjamin and Tedrake, Russ and Song, Shuran},
  journal={The International Journal of Robotics Research},
  pages={02783649241273668},
  year={2023},
  publisher={SAGE Publications Sage UK: London, England}
}

@article{liu2024libero,
  title={Libero: Benchmarking knowledge transfer for lifelong robot learning},
  author={Liu, Bo and Zhu, Yifeng and Gao, Chongkai and Feng, Yihao and Liu, Qiang and Zhu, Yuke and Stone, Peter},
  journal={Advances in Neural Information Processing Systems},
  volume={36},
  year={2024}
}

@article{black2024pi_0,
  title={pi0: A Vision-Language-Action Flow Model for General Robot Control},
  author={Black, Kevin and Brown, Noah and Driess, Danny and Esmail, Adnan and Equi, Michael and Finn, Chelsea and Fusai, Niccolo and Groom, Lachy and Hausman, Karol and Ichter, Brian and others},
  journal={arXiv preprint arXiv:2410.24164},
  year={2024}
}

@article{lipman2022flow,
  title={Flow matching for generative modeling},
  author={Lipman, Yaron and Chen, Ricky TQ and Ben-Hamu, Heli and Nickel, Maximilian and Le, Matt},
  journal={arXiv preprint arXiv:2210.02747},
  year={2022}
}

@article{karamcheti2024prismatic,
  title={Prismatic vlms: Investigating the design space of visually-conditioned language models},
  author={Karamcheti, Siddharth and Nair, Suraj and Balakrishna, Ashwin and Liang, Percy and Kollar, Thomas and Sadigh, Dorsa},
  journal={arXiv preprint arXiv:2402.07865},
  year={2024}
}

@article{touvron2023llama,
  title={Llama 2: Open foundation and fine-tuned chat models},
  author={Touvron, Hugo and Martin, Louis and Stone, Kevin and Albert, Peter and Almahairi, Amjad and Babaei, Yasmine and Bashlykov, Nikolay and Batra, Soumya and Bhargava, Prajjwal and Bhosale, Shruti and others},
  journal={arXiv preprint arXiv:2307.09288},
  year={2023}
}

@article{kim2025fine,
  title={Fine-tuning vision-language-action models: Optimizing speed and success},
  author={Kim, Moo Jin and Finn, Chelsea and Liang, Percy},
  journal={arXiv preprint arXiv:2502.19645},
  year={2025}
}

@article{li2025cogvla,
  title={CogVLA: Cognition-Aligned Vision-Language-Action Model via Instruction-Driven Routing \& Sparsification},
  author={Li, Wei and Zhang, Renshan and Shao, Rui and He, Jie and Nie, Liqiang},
  journal={arXiv preprint arXiv:2508.21046},
  year={2025}
}

@article{intelligence2025pi,
  title={{$\pi_{0.5}$: a Vision-Language-Action Model with Open-World Generalization}},
  author={Intelligence, Physical and Black, Kevin and Brown, Noah and Darpinian, James and Dhabalia, Karan and Driess, Danny and Esmail, Adnan and Equi, Michael and Finn, Chelsea and Fusai, Niccolo and others},
  journal={arXiv preprint arXiv:2504.16054},
  year={2025}
}

@article{li2025unified,
  title={Unified video action model},
  author={Li, Shuang and Gao, Yihuai and Sadigh, Dorsa and Song, Shuran},
  journal={arXiv preprint arXiv:2503.00200},
  year={2025}
}

@article{bu2025univla,
  title={Univla: Learning to act anywhere with task-centric latent actions},
  author={Bu, Qingwen and Yang, Yanting and Cai, Jisong and Gao, Shenyuan and Ren, Guanghui and Yao, Maoqing and Luo, Ping and Li, Hongyang},
  journal={arXiv preprint arXiv:2505.06111},
  year={2025}
}

@article{liang2025video,
  title={Video Generators are Robot Policies},
  author={Liang, Junbang and Tokmakov, Pavel and Liu, Ruoshi and Sudhakar, Sruthi and Shah, Paarth and Ambrus, Rares and Vondrick, Carl},
  journal={arXiv preprint arXiv:2508.00795},
  year={2025}
}

@misc{nvidia2025cosmosworldfoundationmodel,
      title={Cosmos World Foundation Model Platform for Physical AI}, 
      author={NVIDIA and : and Niket Agarwal and Arslan Ali and Maciej Bala and Yogesh Balaji and Erik Barker and Tiffany Cai and Prithvijit Chattopadhyay and Yongxin Chen and Yin Cui and Yifan Ding and Daniel Dworakowski and Jiaojiao Fan and Michele Fenzi and Francesco Ferroni and Sanja Fidler and Dieter Fox and Songwei Ge and Yunhao Ge and Jinwei Gu and Siddharth Gururani and Ethan He and Jiahui Huang and Jacob Huffman and Pooya Jannaty and Jingyi Jin and Seung Wook Kim and Gergely Klár and Grace Lam and Shiyi Lan and Laura Leal-Taixe and Anqi Li and Zhaoshuo Li and Chen-Hsuan Lin and Tsung-Yi Lin and Huan Ling and Ming-Yu Liu and Xian Liu and Alice Luo and Qianli Ma and Hanzi Mao and Kaichun Mo and Arsalan Mousavian and Seungjun Nah and Sriharsha Niverty and David Page and Despoina Paschalidou and Zeeshan Patel and Lindsey Pavao and Morteza Ramezanali and Fitsum Reda and Xiaowei Ren and Vasanth Rao Naik Sabavat and Ed Schmerling and Stella Shi and Bartosz Stefaniak and Shitao Tang and Lyne Tchapmi and Przemek Tredak and Wei-Cheng Tseng and Jibin Varghese and Hao Wang and Haoxiang Wang and Heng Wang and Ting-Chun Wang and Fangyin Wei and Xinyue Wei and Jay Zhangjie Wu and Jiashu Xu and Wei Yang and Lin Yen-Chen and Xiaohui Zeng and Yu Zeng and Jing Zhang and Qinsheng Zhang and Yuxuan Zhang and Qingqing Zhao and Artur Zolkowski},
      year={2025},
      eprint={2501.03575},
      archivePrefix={arXiv},
      primaryClass={cs.CV},
      url={https://arxiv.org/abs/2501.03575}, 
}

@article{wan2025,
      title={Wan: Open and Advanced Large-Scale Video Generative Models}, 
      author={Team Wan and Ang Wang and Baole Ai and Bin Wen and Chaojie Mao and Chen-Wei Xie and Di Chen and Feiwu Yu and Haiming Zhao and Jianxiao Yang and Jianyuan Zeng and Jiayu Wang and Jingfeng Zhang and Jingren Zhou and Jinkai Wang and Jixuan Chen and Kai Zhu and Kang Zhao and Keyu Yan and Lianghua Huang and Mengyang Feng and Ningyi Zhang and Pandeng Li and Pingyu Wu and Ruihang Chu and Ruili Feng and Shiwei Zhang and Siyang Sun and Tao Fang and Tianxing Wang and Tianyi Gui and Tingyu Weng and Tong Shen and Wei Lin and Wei Wang and Wei Wang and Wenmeng Zhou and Wente Wang and Wenting Shen and Wenyuan Yu and Xianzhong Shi and Xiaoming Huang and Xin Xu and Yan Kou and Yangyu Lv and Yifei Li and Yijing Liu and Yiming Wang and Yingya Zhang and Yitong Huang and Yong Li and You Wu and Yu Liu and Yulin Pan and Yun Zheng and Yuntao Hong and Yupeng Shi and Yutong Feng and Zeyinzi Jiang and Zhen Han and Zhi-Fan Wu and Ziyu Liu},
      journal = {arXiv preprint arXiv:2503.20314},
      year={2025}
}

@inproceedings{peebles2023scalable,
  title={Scalable diffusion models with transformers},
  author={Peebles, William and Xie, Saining},
  booktitle={Proceedings of the IEEE/CVF international conference on computer vision},
  pages={4195--4205},
  year={2023}
}

@article{hou2025dita,
  title={Dita: Scaling diffusion transformer for generalist vision-language-action policy},
  author={Hou, Zhi and Zhang, Tianyi and Xiong, Yuwen and Duan, Haonan and Pu, Hengjun and Tong, Ronglei and Zhao, Chengyang and Zhu, Xizhou and Qiao, Yu and Dai, Jifeng and others},
  journal={arXiv preprint arXiv:2503.19757},
  year={2025}
}

@article{zhong2025flowvla,
  title={FlowVLA: Thinking in Motion with a Visual Chain of Thought},
  author={Zhong, Zhide and Yan, Haodong and Li, Junfeng and Liu, Xiangchen and Gong, Xin and Song, Wenxuan and Chen, Jiayi and Li, Haoang},
  journal={arXiv preprint arXiv:2508.18269},
  year={2025}
}

@article{hu2024video,
  title={Video prediction policy: A generalist robot policy with predictive visual representations},
  author={Hu, Yucheng and Guo, Yanjiang and Wang, Pengchao and Chen, Xiaoyu and Wang, Yen-Jen and Zhang, Jianke and Sreenath, Koushil and Lu, Chaochao and Chen, Jianyu},
  journal={arXiv preprint arXiv:2412.14803},
  year={2024}
}

@article{liao2025genie,
  title={Genie envisioner: A unified world foundation platform for robotic manipulation},
  author={Liao, Yue and Zhou, Pengfei and Huang, Siyuan and Yang, Donglin and Chen, Shengcong and Jiang, Yuxin and Hu, Yue and Cai, Jingbin and Liu, Si and Luo, Jianlan and others},
  journal={arXiv preprint arXiv:2508.05635},
  year={2025}
}

@misc{unifolm-wma-0,
  author       = {Unitree},
  title        = {UnifoLM-WMA-0: A World-Model-Action (WMA) Framework under UnifoLM Family},
  year         = {2025},
}

@article{feng2025vidar,
  title={Vidar: Embodied Video Diffusion Model for Generalist Bimanual Manipulation},
  author={Feng, Yao and Tan, Hengkai and Mao, Xinyi and Liu, Guodong and Huang, Shuhe and Xiang, Chendong and Su, Hang and Zhu, Jun},
  journal={arXiv preprint arXiv:2507.12898},
  year={2025}
}

@article{wang2025latent,
  title={Latent Policy Steering with Embodiment-Agnostic Pretrained World Models},
  author={Wang, Yiqi and Verghese, Mrinal and Schneider, Jeff},
  journal={arXiv preprint arXiv:2507.13340},
  year={2025}
}

@article{zheng2025flare,
  title={FLARE: Robot learning with implicit world modeling},
  author={Zheng, Ruijie and Wang, Jing and Reed, Scott and Bjorck, Johan and Fang, Yu and Hu, Fengyuan and Jang, Joel and Kundalia, Kaushil and Lin, Zongyu and Magne, Loic and others},
  journal={arXiv preprint arXiv:2505.15659},
  year={2025}
}

@article{kong2024hunyuanvideo,
  title={Hunyuanvideo: A systematic framework for large video generative models},
  author={Kong, Weijie and Tian, Qi and Zhang, Zijian and Min, Rox and Dai, Zuozhuo and Zhou, Jin and Xiong, Jiangfeng and Li, Xin and Wu, Bo and Zhang, Jianwei and others},
  journal={arXiv preprint arXiv:2412.03603},
  year={2024}
}

@article{zheng2024open,
  title={Open-sora: Democratizing efficient video production for all},
  author={Zheng, Zangwei and Peng, Xiangyu and Yang, Tianji and Shen, Chenhui and Li, Shenggui and Liu, Hongxin and Zhou, Yukun and Li, Tianyi and You, Yang},
  journal={arXiv preprint arXiv:2412.20404},
  year={2024}
}

@article{bjorck2025gr00t,
  title={Gr00t n1: An open foundation model for generalist humanoid robots},
  author={Bjorck, Johan and Casta{\~n}eda, Fernando and Cherniadev, Nikita and Da, Xingye and Ding, Runyu and Fan, Linxi and Fang, Yu and Fox, Dieter and Hu, Fengyuan and Huang, Spencer and others},
  journal={arXiv preprint arXiv:2503.14734},
  year={2025}
}

@article{nasiriany2024robocasa,
  title={Robocasa: Large-scale simulation of everyday tasks for generalist robots},
  author={Nasiriany, Soroush and Maddukuri, Abhiram and Zhang, Lance and Parikh, Adeet and Lo, Aaron and Joshi, Abhishek and Mandlekar, Ajay and Zhu, Yuke},
  journal={arXiv preprint arXiv:2406.02523},
  year={2024}
}

@article{zhao2025xrobotoolkit,
      title={XRoboToolkit: A Cross-Platform Framework for Robot Teleoperation}, 
      author={Zhigen Zhao and Liuchuan Yu and Ke Jing and Ning Yang}, 
      journal={arXiv preprint arXiv:2508.00097},
      year={2025}
}

@article{aldaco2024aloha,
  title={Aloha 2: An enhanced low-cost hardware for bimanual teleoperation},
  author={Aldaco, Jorge and Armstrong, Travis and Baruch, Robert and Bingham, Jeff and Chan, Sanky and Draper, Kenneth and Dwibedi, Debidatta and Finn, Chelsea and Florence, Pete and Goodrich, Spencer and others},
  journal={arXiv preprint arXiv:2405.02292},
  year={2024}
}
\bibliographystyle{main_conference}

\newpage
\appendix
\section{Appendix}

\subsection{Model \& Training Configurations}

\begin{table}[h]
\centering
\small
\begin{tabular}{lc}
\toprule
\textbf{Parameter} & \textbf{Value} \\
\midrule
\multicolumn{2}{c}{\textit{Video DiT}} \\
\midrule
Base VGM               & Cosmos-Predict2.5-2B~\citep{cosmos25} \\
Attention Implementation & flash\_attention\_2 \\
Hidden Feature Dim          & 2048 \\
Extract Layer               & 18 \\
\midrule
\multicolumn{2}{c}{\textit{Action DiT}} \\
\midrule
Action Model Type          & DiT-B \\
Hidden Size                & 2560 \\
Add Positional Embedding   & True \\
Max Sequence Length         & 1024 \\
Action Dim                 & 32 \\
State Dim                  & 64 \\
Future Action Window Size  & 15 \\
Action Horizon             & 16 \\
Past Action Window Size    & 0 \\
Cross Attention Dim           & 2048 \\
Dropout                       & 0.2 \\
Final Dropout                 & True \\
Interleave Self Attention     & True \\
Norm Type                     & AdaLN \\
Num Layers                    & 16 \\
Output Dim                    & 2560 \\
Repeated Diffusion Steps (train)  & 4 \\
Noise $\beta$ ($\alpha$)          & 1.5 \\
Noise $\beta$ ($\beta$)           & 1.0 \\
Noise $s$                         & 0.999 \\
Num Timestep Buckets              & 1000 \\
Num Inference Timesteps            & 4 \\
\midrule
\multicolumn{2}{c}{\textit{Training Configurations}} \\
\midrule
Per Device Batch Size & 8 \\
Num of GPUs &32 \\
Max Train Steps              & 100000 \\
Num Warmup Steps             & 5000 \\
Warmup Ratio                 & 0.1 \\
VGM LR             & $1 \times 10^{-5}$ \\
Action Model LR              & $1 \times 10^{-4}$ \\
LR Scheduler                 & cosine\_with\_min\_lr \\
Min LR                       & $5 \times 10^{-7}$ \\
Gradient Clipping            & 1.0 \\
Gradient Accumulation Steps  & 1 \\
Optimizer        & AdamW \\
$\beta_1, \beta_2$ & $(0.9, 0.95)$ \\
$\epsilon$       & $1 \times 10^{-8}$ \\
Weight Decay     & $1 \times 10^{-8}$ \\
\bottomrule
\end{tabular}
\caption{Model \& training configurations}
\end{table}

\subsection{Dataset Configuration}
We detail the composition and configuration of the datasets used to train and evaluate the DiT4DiT framework, as summarized in Table~\ref{tab:dataset_statistics}. To rigorously assess both fundamental learning capabilities and physical deployability, our dataset usage is strategically partitioned into two distinct pipelines: one for simulated benchmark evaluation and another for real-world deployment.

\textbf{Simulated benchmark data}: To evaluate our framework in simulated environments, we train the models directly on the target datasets. For the RoboCasa-GR1~\citep{nasiriany2024robocasa} tabletop tasks, we utilize the Fourier\_GR1\_Unified\_1K dataset~\citep{bjorck2025gr00t}, which consists of 24,000 demonstration episodes collected using the highly complex 29-DoF GR1 humanoid embodiment. For the LIBERO~\citep{liu2024libero} benchmark, we use its official dataset comprising 1,693 episodes based on a 7-DoF Franka Emika Panda robotic arm. Training from scratch on these datasets ensures a strictly fair evaluation against baseline methods in simulation as we do not have access to all pre-training datasets of various methods.

\textbf{Pre-training data for real-world tasks}: To facilitate robust physical deployment, DiT4DiT undergoes a crucial pre-training phase to acquire fundamental spatiotemporal and physical priors. For this stage, we utilize the scaled Fourier\_GR1\_Pretrain\_10K dataset~\citep{bjorck2025gr00t}, comprising 241,450 episodes of the 29-DoF GR1 embodiment. As highlighted in our main text, this pre-training corpus represents merely \textbf{15\%} of the massive data volume leveraged by baselines like GR00T-N1.5, heavily emphasizing the data efficiency of our generative video backbone.

\textbf{Real-world fine-tuning data}: Following pre-training, the model is fine-tuned to adapt its generative priors to the target physical robot. For this stage, we employ our custom real-robot dataset, consisting of 1,400 high-quality, teleoperated demonstration episodes (200 episodes for each task). This dataset is specifically tailored for the Unitree G1 humanoid robot, operating with a continuous 16-DoF action space. This crucial fine-tuning phase successfully grounds the broad simulation-based physical dynamics into precise, real-world continuous control commands.

\begin{table}[htbp]
\centering
\small
\begin{tabular}{lrlr}
\toprule
Dataset & Episode Cnt & Embodiment & DoF \\
\midrule
Fourier\_GR1\_Unified\_1K~\citep{bjorck2025gr00t} & 24,000  & GR1 humanoid   & 29 \\
Fourier\_GR1\_Pretrain\_10K~\citep{bjorck2025gr00t} & 241,450 & GR1 humanoid   & 29 \\
LIBERO~\citep{liu2024libero}      & 1,693   & Franka Emika Panda & 7  \\
Real Robot                 & 1,400   & G1 humanoid    & 16 \\
\bottomrule
\end{tabular}
\caption{\textbf{Details of the used datasets}. We report the episode count, embodiment type, and degrees of freedom.}
\label{tab:dataset_statistics}
\end{table}

\subsection{Real-world Experiment Setting}
\begin{figure}
    \centering
    \includegraphics[width=0.6\linewidth]{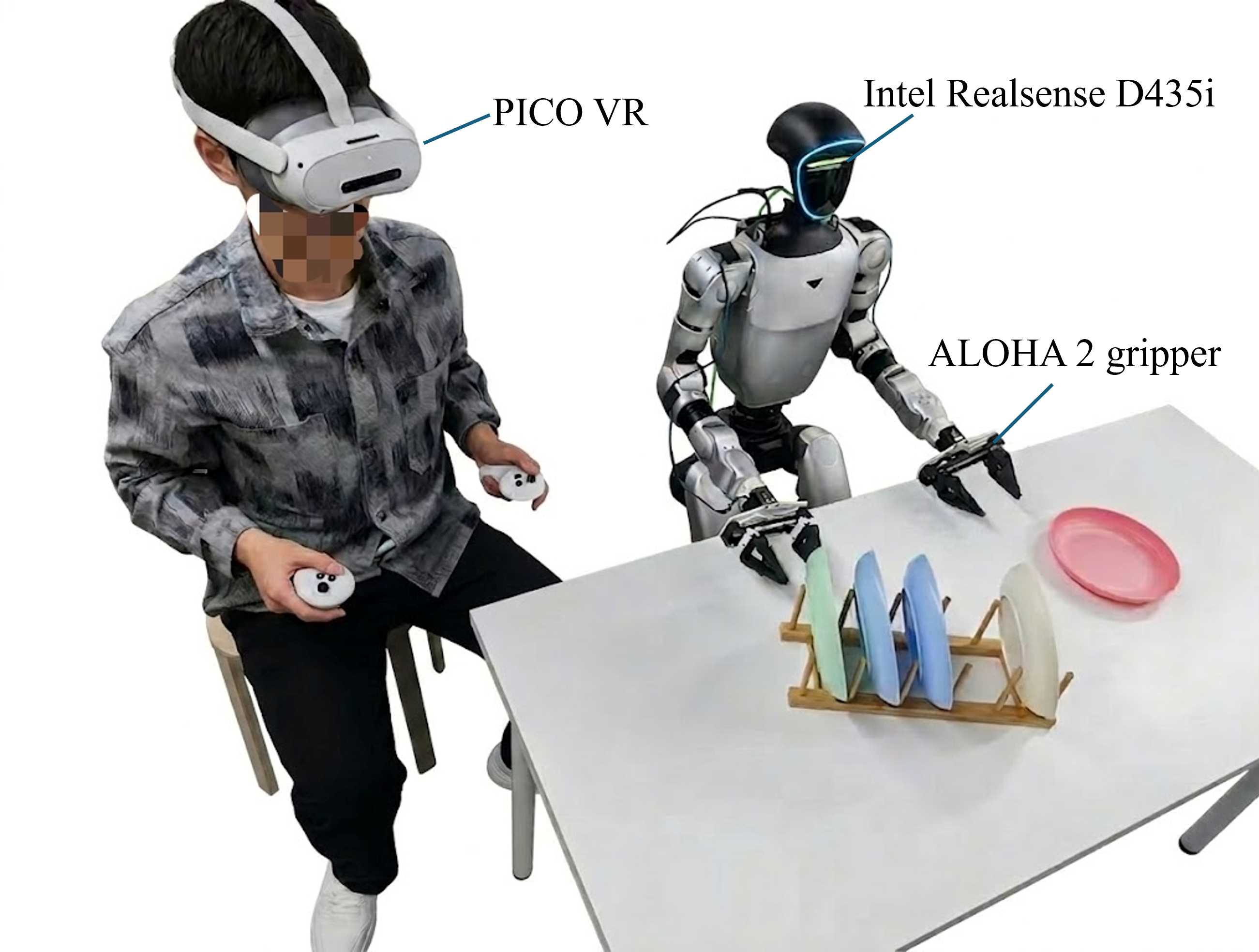}
    \caption{\textbf{Visualization of the robot system setups.} The experimental platform consists of a Unitree G1 humanoid robot equipped with dual ALOHA 2~\citep{aldaco2024aloha} grippers and an Intel RealSense D435i camera. A human operator utilizes a PICO VR headset and the XRoboToolkit~\citep{zhao2025xrobotoolkit} framework to perform teleoperated demonstrations for high-quality data collection.}
    \label{fig:real-setting}
\end{figure}
As shown in fig.~\ref{fig:real-setting}, our real-world experimental system is built upon the Unitree G1 humanoid robot, featuring a 16-DoF action space driven by dual 7-DoF arms. Each arm is equipped with an ALOHA 2~\citep{aldaco2024aloha} gripper to facilitate high-precision bimanual manipulation. For visual perception, an Intel RealSense D435i camera is mounted on the robot’s head to capture ego-centric RGB observations at a resolution of 640x480. The real-time inference is executed on a workstation with a single NVIDIA GeForce RTX 4090 GPU.

Data collection is performed via a VR-based teleoperation pipeline using a PICO VR headset and handheld controllers. This setup allows a human operator to provide natural demonstrations by mapping motion directly to the robot's joints. The XRoboToolkit~\citep{zhao2025xrobotoolkit} framework manages the entire software stack, ensuring precise synchronization of multimodal sensor data and high-quality recording of demonstration trajectories. This integrated environment supports a seamless transition from human-led data collection to autonomous policy deployment.

\subsection{Limitations and Discussion}

While DiT4DiT demonstrates exceptional capabilities in bridging generative video priors and continuous robotic control, limitations present exciting avenues for future research.

Our current physical deployment relies exclusively on a single egocentric (ego-view) camera. While this highlights the remarkable spatial reasoning capabilities of our video representations, single-view setups are fundamentally susceptible to severe visual occlusions. In complex bimanual tasks, the robot's own arms or larger objects can temporarily block the camera's line of sight, potentially disrupting the temporal continuity of the visual features. Future iterations could explore the integration of auxiliary sensory inputs, such as wrist-mounted cameras or tactile feedback, fusing these modalities with the video DiT backbone to maintain robust state estimation during severe occlusions.

Our real-world experiments achieved state-of-the-art zero-shot generalization using a pre-training corpus that represents merely 15\% of the data volume utilized by contemporary large-scale models like GR00T. A natural and promising next step is to drastically scale the pre-training data across diverse robotic embodiments (e.g., varying kinematics, grippers, and camera parameters). Given the data-efficient nature of our dual flow-matching objective, scaling DiT4DiT with massive, cross-embodiment datasets could yield a highly generalized robotic foundation model, further solidifying video generation as the optimal scaling proxy for embodied intelligence.

\begin{figure}[htbp]
    \centering
    \includegraphics[width=1.0\linewidth]{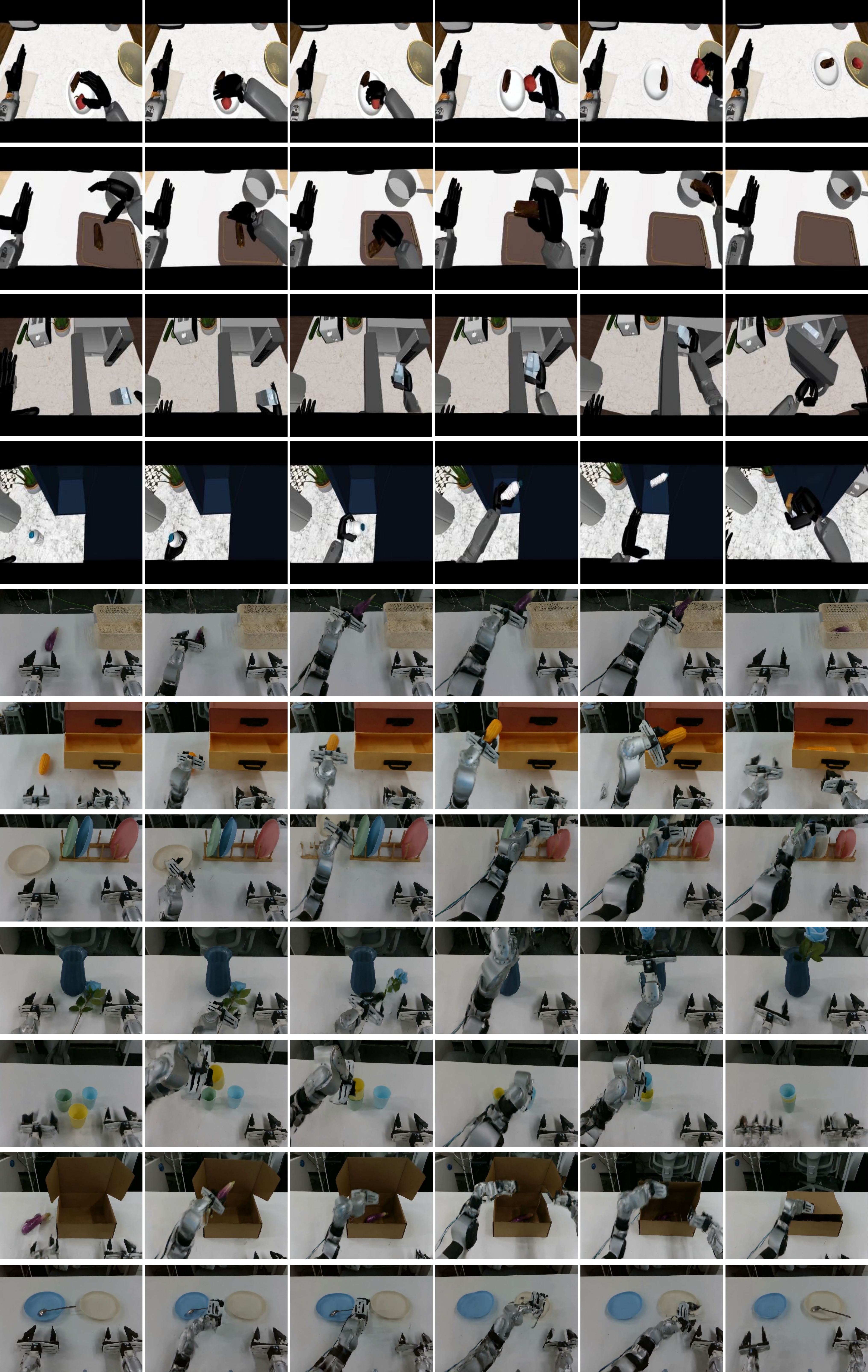}
    \caption{Future video rollouts generated by DiT4DiT. }
    \label{fig:video_gen}
\end{figure}

\end{document}